\documentclass[10pt,twocolumn,letterpaper]{article}

\usepackage{cvpr}
\usepackage{times}
\usepackage{epsfig}
\usepackage{graphicx}
\usepackage{amsmath}
\usepackage{amssymb}
\usepackage{multirow}
\usepackage{multicol}
\usepackage{subcaption}
\usepackage{caption}
\usepackage{afterpage}
\usepackage{float}
\usepackage{enumitem}
\usepackage[pagebackref=true,breaklinks=true,letterpaper=true,colorlinks,bookmarks=false]{hyperref}

\cvprfinalcopy 

\def\cvprPaperID{2958} 

\begin{document}

\title{A Compact Embedding for Facial Expression Similarity}

\author{Raviteja Vemulapalli\\
Google AI\\
{\tt\small ravitejavemu@google.com}
\and
Aseem Agarwala\\
Google AI\\
{\tt\small aseemaa@google.com}
}
\vspace{-10pt}
\maketitle
\begin{abstract}
Most of the existing work on automatic facial expression analysis focuses on discrete emotion recognition, or facial action unit detection. However, facial expressions do not always fall neatly into pre-defined semantic categories. Also, the similarity between expressions measured in the action unit space need not correspond to how humans perceive expression similarity. Different from previous work, our goal is to describe facial expressions in a continuous fashion using a compact embedding space that mimics human visual preferences. To achieve this goal, we collect a large-scale faces-in-the-wild dataset with human annotations in the form: Expressions A and B are visually more similar when compared to expression C, and use this dataset to train a neural network that produces a compact (16-dimensional) expression embedding. We experimentally demonstrate that the learned embedding can be successfully used for various applications such as expression retrieval, photo album summarization, and emotion recognition. We also show that the embedding learned using the proposed dataset performs better than several other embeddings learned using existing emotion or action unit datasets.
\end{abstract}
\vspace{-3mm}
\section{Introduction}
\begin{figure}[t]
    \centering
    \includegraphics[width=0.45\textwidth, height=0.20\textwidth]{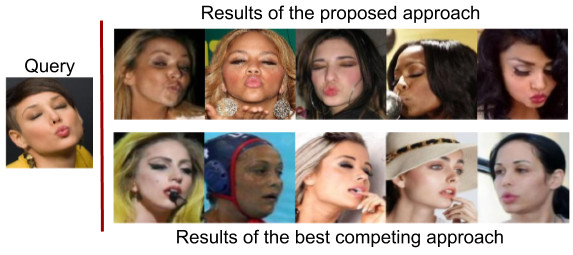}
    \caption{Expression retrieval results for embeddings learned using the proposed dataset (top) and an existing emotion classification dataset (bottom).}
    \label{fig:example}
\end{figure}

Automatic facial expression analysis has received significant attention from the computer vision community due to its numerous applications such as emotion prediction, expression retrieval (Figure~\ref{fig:example}), photo album summarization, candid portrait selection~\cite{Candid}, etc. Most of the existing work~\cite{EmotionSurvey,AUSurvey} focuses on recognizing discrete emotions or action units defined by the Facial Action Coding System (FACS)~\cite{FACS}. However, facial expressions do not always fit neatly into semantic boxes, and there could be significant variations in the expression within the same semantic category. For example, smiles can come in many subtle variations, from shy smiles, to nervous smiles, to laughter. Also, not every human-recognizable facial expression has a name. In general, the space of facial expressions can be viewed as a continuous, multi-dimensional space.

 In this work, we focus on learning a compact, language-free, subject-independent, and continuous expression embedding space that mimics human visual preferences. If humans consider two expressions to be visually more similar when compared to a third one, then the distance between these two expressions in the embedding space should be smaller than their distances from the third expression. To learn such an embedding we collect a new dataset, referred to as the Facial Expression Comparison (FEC) dataset, that consists of around 500K expression triplets generated using 156K face images, along with annotations that specify which two expressions in each triplet are most similar to each other. To the best of our knowledge, this is the first large-scale face dataset with expression comparison annotations. This dataset can be downloaded from \url{https://ai.google/tools/datasets/google-facial-expression/}.

We show that a compact (16-dimensional) expression embedding space can be learned by training a deep network with the proposed FEC dataset using triplet loss~\cite{FaceNet}. Based on the distances in the learned embedding space, we are able to predict the most similar pair in a triplet with an accuracy of 81.8\% when evaluated on a held-out validation set. The accuracy of median human rater is 87.5\% on this validation set, and the accuracy of random selection is 33.3\%. We also show that the embedding learned using the FEC dataset performs better than several other embeddings learned using existing emotion or action unit datasets.

We experimentally demonstrate that the expression embedding learned using the FEC dataset can be successfully used for various applications such as expression retrieval, photo album summarization, and emotion recognition.
\subsection{Our contributions}
\begin{itemize}[topsep=2pt]
\itemsep 0pt
    \item We introduce the FEC dataset, which is the first large-scale face dataset with expression comparison annotations. This dataset is now available to public.
    \item We experimentally demonstrate that a 16-dimensional expression embedding learned by training a deep neural network with the FEC dataset can be successfully used for several expression-based applications.
    \item We show that the embedding learned using the FEC dataset performs better than several other embeddings learned using existing emotion or action unit datasets.
\end{itemize}
\section{Related work}
Most of the existing research in the area of automatic facial expression analysis focuses on the following three topics: \textit{(i) Categorical model:} Assigning discrete emotion category labels, \textit{(ii) FACS model}: Detecting the presence/absence (and the strength) of various action units defined by FACS~\cite{FACS}, and \textit{(iii) Dimensional model}: Describing emotions using two or three dimensional models such as valence-arousal~\cite{ValAro}, pleasure-arousal-dominance~\cite{PAD}, etc.\ Summarizing the vast amount of existing research on these topics is beyond the scope of this paper and we refer the readers to~\cite{Aff-wild, EmotionSurvey,AUSurvey} for recent surveys on these topics.\vspace{10pt}\\
\noindent \textbf{Expression datasets:}\ Several facial expression datasets have been created in the past that consist of face images labeled with discrete emotion categories~\cite{Emotionnet,afew-7,Afew,Sfew,Fer2013,Multipie,Raf-db,CK+,AffectNet,Fer-Wild,MMI,ExpW,Oulu-casia}, facial action units~\cite{Emotionnet,CK+,Disfa,AM-FED,MMI}, and strengths of valence and arousal~\cite{Deap,Aff-wild,afew-va,AffectNet,Reloca}. While these datasets played a significant role in the advancement of automatic facial expression analysis in terms of emotion recognition, action unit detection and valence-arousal estimation, they are not the best fit for learning a compact expression embedding space that mimics human visual preferences.
\vspace{10pt}\\
\noindent \textbf{Expression embedding:}
A neural network was trained in~\cite{TripletEmbedding} using an emotion classification dataset and category label-based triplet loss~\cite{FaceNet} to produce a 128-dimensional embedding, which was combined with an LSTM-based network for animating three basic expressions. Emotion labels do not provide information about within-class variations and hence a network trained with label-based triplets may not encode fine-grained expression information. The proposed FEC dataset addresses this issue by including expression comparison annotations for within-class triplets.

A self-supervised approach was proposed in~\cite{EmbeddingBMVC18} to learn a 256-dimensional facial attribute embedding by watching videos, and the learned embedding was used for multiple tasks such as head pose estimation, facial landmarks prediction, and emotion recognition by training an additional classification or regression layer using labeled training data. However, as reported in~\cite{EmbeddingBMVC18}, its performance is worse than existing approaches on these tasks. Different from~\cite{EmbeddingBMVC18}, we follow a fully-supervised approach for learning a compact (16-dimensional) expression embedding.\vspace{10pt}\\
\noindent \textbf{Triplet loss-based representation learning:}
Several existing works have used triplet-based loss functions for learning image representations. While majority of them use category label-based triplets~\cite{HTL,TCLoss,PersonReId,DRDL,FaceNet,LiftingLoss,AngularLoss,BinaryEmb}, some existing works~\cite{ImgSimJMLR,DeepRanking} have focused on learning fine-grained representations. While~\cite{DeepRanking} used a similarity measure computed using several existing feature representations to generate groundtruth annotations for the triplets,~\cite{ImgSimJMLR} used text-image relevance based on Google image search to annotate the triplets. Different from these approaches, we use human raters to annotate the triplets. Also, none of these works focus on facial expressions.

\begin{figure*}[t]
\centering
\begin{minipage}[t]{0.3\textwidth}
\centering
\begin{minipage}[t]{0.3\textwidth}
\includegraphics[width=\textwidth, height=\textwidth]{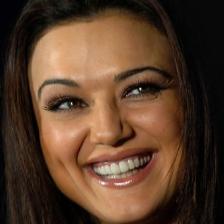}
\caption*{Happiness}
\end{minipage}
\begin{minipage}[t]{0.3\textwidth}
\includegraphics[width=\textwidth, height=\textwidth]{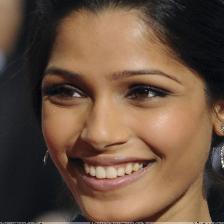}
\caption*{Happiness}
\end{minipage}
\begin{minipage}[t]{0.3\textwidth}
\includegraphics[width=\textwidth, height=\textwidth]{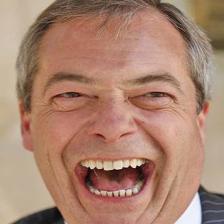}
\caption*{Happiness}
\end{minipage}
\caption*{(a) One-class triplet}
\end{minipage}
\vspace{5pt}
\begin{minipage}[t]{0.3\textwidth}
\centering
\begin{minipage}[t]{0.3\textwidth}
\includegraphics[width=\textwidth, height=\textwidth]{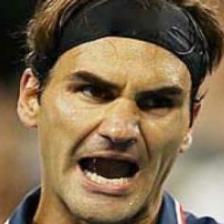}
\caption*{Anger}
\end{minipage}
\begin{minipage}[t]{0.3\textwidth}
\includegraphics[width=\textwidth, height=\textwidth]{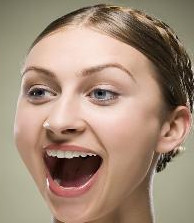}
\caption*{Surprise}
\end{minipage}
\begin{minipage}[t]{0.3\textwidth}
\includegraphics[width=\textwidth, height=\textwidth]{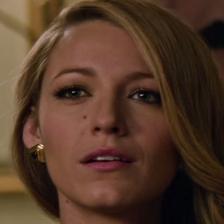}
\caption*{Surprise}
\end{minipage}
\caption*{(b) Two-class triplet}
\end{minipage}
\vspace{5pt}
\begin{minipage}[t]{0.3\textwidth}
\centering
\begin{minipage}[t]{0.3\textwidth}
\includegraphics[width=\textwidth, height=\textwidth]{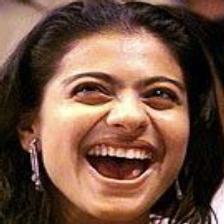}
\caption*{Happiness}
\end{minipage}
\begin{minipage}[t]{0.3\textwidth}
\includegraphics[width=\textwidth, height=\textwidth]{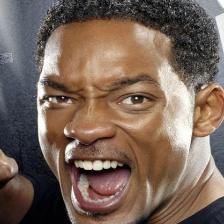}
\caption*{Anger}
\end{minipage}
\begin{minipage}[t]{0.3\textwidth}
\includegraphics[width=\textwidth, height=\textwidth]{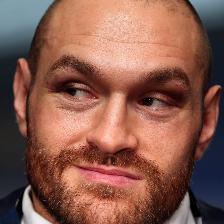}
\caption*{Contempt}
\end{minipage}
\caption*{(c) Three-class triplet}
\end{minipage}
\vspace{-3mm}
\caption{Different types of triplets based on the emotion labels used in the AffectNet~\cite{AffectNet} dataset.}
\label{fig:triplets}
\end{figure*}
\section{Facial expression comparison dataset}
\label{sec::motivation}
In this section, we introduce the FEC dataset, which is a large-scale faces-in-the-wild dataset with expression comparison annotations provided by human raters. To the best of our knowledge, there is no such publicly-available expression comparison dataset. Most of the existing expression datasets are either annotated with emotion labels, or facial action units, or strengths of valence and arousal. 

One may think that we could generate comparison annotations for the existing datasets using the available emotion or action unit labels. However, there are several issues with such an approach:
\begin{itemize}[topsep=3pt]
\itemsep 0pt
\item[$\bullet$] Emotion labels do not provide information about within-class variations and hence we cannot generate comparison annotations within a class. For example, while all the expressions in Figure~\ref{fig:triplets}(a) fall into the \textit{Happiness} category, the left and middle expressions are visually more similar when compared to the right expression. Such within-class comparisons are important to learn a fine-grained expression representation.
\item[$\bullet$] Due to within-class variations and between-class similarities, two expressions from the same category need not be visually more similar when compared to an expression from a different category. For example, while the middle and right expressions in Figure~\ref{fig:triplets}(b) belong to the \textit{Surprise} category, the middle expression is visually more similar to the left expression which belongs to the \textit{Anger} category.
\item[$\bullet$] It is difficult to predict the visual similarity relationships between expressions from three different emotion categories by using labels. For example, while the three expressions in Figure~\ref{fig:triplets}(c) belong to three different categories, the left and middle expressions are visually more similar when compared to the right expression. Such comparisons are useful for learning an embedding that can model long-range visual similarity relationships between different categories.
\item[$\bullet$] It is unclear how the difference in the strengths of action units between two expressions could be converted into a distance function that mimics visual preferences. 
\end{itemize}

\begin{figure*}
\centering
\includegraphics[width=\textwidth]{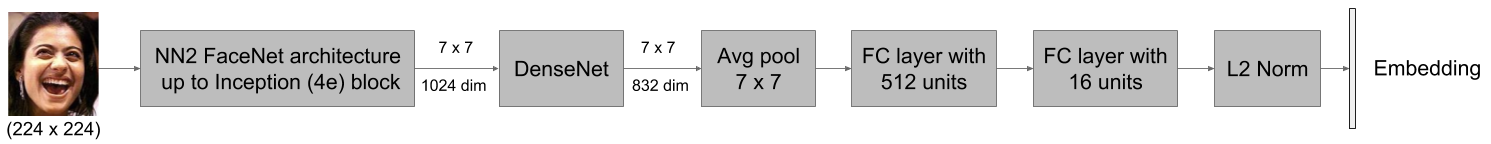}
\caption{Proposed embedding network based on the NN2 FaceNet architecture~\cite{FaceNet}. Here, FC stands for fully-connected, the L2-Norm layer performs $\ell_2$ normalization, and DenseNet consists of a $1\times 1$ convolution layer (512 filters) followed by a Dense block~\cite{DenseNet} with 5 layers and growth rate of 64.}
\label{fig:network}
\end{figure*}
\subsection{Dataset}
\label{sec::dataset}
Each sample in the FEC dataset consists of a face image triplet $(I_1, I_2, I_3)$ along with a label $L \in \{1, 2, 3\}$ that indicates which two images in the triplet form the most similar pair in terms of facial expression. For example, $L=1$ means $I_2$ and $I_3$ are visually more similar when compared to $I_1$. Note that these triplets do not have a notion of anchor, and each triplet provides two annotations: $I_2$ is closer to $I_3$ than $I_1$, and $I_3$ is closer to $I_2$ than $I_1$. This is different from the commonly-used triplet annotation~\cite{FaceNet,LMNN} that consists of an anchor, a positive image and a negative image. Also, in this dataset, an image A can be (relatively) closer to another image B in one triplet and (relatively) farther from the same image B in another triplet. This is different from the triplets generated using category labels~\cite{FaceNet}, in which any two images will either form a similar pair or a dissimilar pair in all the triplets they appear in.
\begin{table*}[t]
    \centering
    \small
    \begin{tabular}{|c|c|c|c|c|c|c|}
    \hline
        \multirow{2}{*}{Partition} & \multirow{2}{*}{Rater agreement} & \multicolumn{4}{|c|}{Triplet type} & \multirow{2}{*}{Faces}\\\cline{3-6}
         & & One-class & Two-class & Three-class & All & \\\hline
        \multirow{3}{*}{Training set} & Strong & 115,544 & 124,665 & 117,540 & 357,749 & \multirow{3}{*}{130,516}\\\cline{2-6}
        & Strong + Weak & 137,266 & 138,034 & 132,435 & 407,735 & \\\cline{2-6}
        & All & 152,674 & 150,234 & 146,235 & 449,143 & \\\hline
        \multirow{3}{*}{Test set} & Strong & 13,046 & 14,607 & 13,941 & 41,594 & \multirow{3}{*}{25,427}\\\cline{2-6}
        & Strong + Weak & 15,411 & 15,908 & 15,404 & 46,723 & \\\cline{2-6}
        & All & 17,059 & 17,107 & 16,894 & 51,060 & \\\hline
        \multirow{3}{*}{Full dataset} & Strong & 128,590 & 139,272 & 131,481 & 399,343 & \multirow{3}{*}{155,943}\\\cline{2-6}
         & Strong + Weak & 152,677 & 153,942 & 147,839 & 454,458 & \\\cline{2-6}
         & All & 169,733 & 167,341 & 163,129 & 500,203 & \\\hline
    \end{tabular}
    \caption{Number of triplets and faces in the proposed FEC dataset.}
    \label{tab:dataset}
\end{table*}

The triplets in the FEC dataset were generated by sampling images from a partially-labeled~\footnote{The images in this dataset are not exhaustively labeled, i.e., an image may not have all the labels that are applicable to it.} internal face dataset in which each face image has one or more of the following emotion labels~\cite{emotions-pnas,emotions-cognitive}: \textit{Amusement, Anger, Awe, Boredom, Concentration, Confusion, Contemplation, Contempt, Contentment, Desire, Disappointment, Disgust, Distress, Doubt, Ecstasy, Elation, Embarrassment, Fear, Interest, Love, Neutral, Pain, Pride, Realization, Relief, Sadness, Shame, Surprise, Sympathy, and Triumph}. To reduce the effect of category-bias, we sampled the images such that all these categories are (roughly) equally represented in the triplet dataset. Each triplet was annotated by six human raters, and the raters were instructed to focus only on expressions ignoring other factors such as identity, gender, ethnicity, pose and age. A total of 40 raters participated in the process, each annotating a subset of the entire dataset.

Based on the existing emotion labels, each triplet in this dataset can be categorized into one of the following types~\footnote{The images in the dataset (from which we sampled the faces) were not exhaustively labeled, and hence, a triplet classified as a two/three-class triplet based on the existing labels may not be be a two/three-class triplet if the images had been exhaustively labeled.}:
\begin{itemize}[topsep=3pt]
\itemsep 0pt
\item[$\bullet$] \textit{One-class triplets}: All the three images share a category label, see Figure~\ref{fig:triplets}(a). These triplets are useful for learning a fine-grained expression representation. 
\item[$\bullet$] \textit{Two-class triplets}: Only two images share a category label and the third image belongs to a different category, see Figure~\ref{fig:triplets}(b). As explained in Section~\ref{sec::motivation}, images sharing a category label need not form the (visually) most similar pair in these triplets.
\item[$\bullet$] \textit{Three-class triplets}: None of the images share a common category label, see Figure~\ref{fig:triplets}(c). These triplets are useful for learning long-range visual similarity relationships between different categories. 
\end{itemize}
One-class triplets are relatively the most difficult ones since the expressions could be very close to each other, and two-class triplets are relatively the easiest ones since the images sharing a label could potentially be different from the remaining image (though not always). While there are other possible types of triplets based on other label combinations (for example, $I_1$, $I_2$ sharing a label, and $I_2$, $I_3$ sharing another label), we prioritized the above three types while collecting the dataset as the other types could be confusing for the raters. Extending the dataset to include the other types will be considered in the future. Table~\ref{tab:dataset} shows the number of triplets in this dataset along with the number of faces used to generate the triplets. The dataset is further divided into training (90\%) and test (10\%) sets, and we encourage the users of this dataset to use the training set for training their algorithms and the test set to validate them.\vspace{10pt}\\
\noindent \textbf{Annotation agreement:}
Each triplet in this dataset was annotated by six raters. For a triplet, we say that the raters \textit{agree strongly} if at least two-thirds of them voted for the maximum-voted label, and \textit{agree weakly} if there is a unique maximum-voted label and half of the raters voted for it. The number of such triplets for each type are shown in Table~\ref{tab:dataset}. Raters agree strongly for about 80\% of the triplets suggesting that humans have a well-defined notion of visual expression similarity.
\section{Facial expression embedding network}
\label{sec:network}
In the recent past, the performance of face recognition systems has improved significantly~\cite{MegaFaceLeaderboard,MSCeleb1MLeaderboard,MegaFace1,MegaFace2} in part due to the availability of large-scale (several million data samples) training datasets such as MS-Celeb-1M~\cite{MSCeleb1M}, MegaFace~\cite{MegaFace2}, SFC~\cite{DeepFace} and Google-Face~\cite{FaceNet}. Neural networks trained on these large-scale datasets see images with significant variations along different dimensions such as lighting, pose, age, gender, ethnicity, etc.\ during training. 

Compared to these large-scale face datasets, our facial expression comparison dataset is significantly smaller (just 130K training faces).\ Hence, in order to leverage the power of a large training set, we build our facial expression embedding network using the pre-trained FaceNet proposed in~\cite{FaceNet}, see Figure~\ref{fig:network}. We use the NN2 version of pre-trained FaceNet~\cite{FaceNet} up to the inception (4e) block~\footnote{We also experimented with features from inception 4d, 5a and 5b blocks, and features from 4e block performed the best.} whose output is a $7 \times 7$ feature map with 1024 channels. This feature map is processed by a DenseNet which consists of a $1\times 1$ convolution layer (512 filters) followed by a Dense block~\cite{DenseNet} with 5 layers and growth rate of 64. The output of DenseNet is passed to a $7\times 7$ average pooling layer followed by a fully connected (FC) layer with 512 hidden units and an embedding layer (a linear FC layer + $\ell_2$ normalization layer). Batch normalization~\cite{BatchNorm} and ReLu6~\cite{Relu6} activation function are used in the DenseNet and the first FC layer. We also use dropout for regularization.

The input to our network is an aligned (rotated to undo roll transformation and scaled to maintain an inter-ocular distance of 55 pixels) $224\times 224$ face image $I$, and the output is a $d$-dimensional embedding $e_{I}$ of unit $\ell_2$ norm.

\subsection{Triplet loss function}
For training the embedding network using the proposed FEC dataset, we use a triplet loss function that encourages the distance between the two images that form the most similar pair to be smaller than the distances of these two images from the third image. For a triplet $(I_1, I_2, I_3)$ with the most similar pair $(I_1, I_2)$, the loss function is given by
\begin{equation}
\begin{aligned}
l(I_1, I_2, I_3) &= max(0, \|e_{I_1} - e_{I_2} \|_2^2 - \|e_{I_1} - e_{I_3} \|_2^2 + \delta)\\
&+ max(0, \|e_{I_1} - e_{I_2} \|_2^2 - \|e_{I_2} - e_{I_3} \|_2^2 + \delta),
\end{aligned}
\end{equation}
where $\delta$ is a small margin.
\section{Experiments}
In this section, we demonstrate the usefulness of the expression embedding learned from the proposed FEC dataset for various applications such as expression retrieval, photo album summarization, and emotion classification. In all our experiments, we only use the triplets with strong rater agreement for both training and evaluation. We also tried using the triplets with weak rater agreement for training, but the results did not improve (see Section~\ref{sec::comparison}). In the rest of the paper, we refer to the proposed expression embedding network trained on the proposed FEC dataset as \textit{FECNet}.
\subsection{Comparative approaches}
Most of the existing large-scale expression datasets focus on the task of classification. One can train a classification network with such a dataset, and use the output of the final or penultimate layer as an expression embedding. Here, we train two networks: \textit{AFFNet-CL} for emotion recognition using the AffectNet dataset~\cite{AffectNet}, and \textit{FACSNet-CL} for facial action unit detection using the DISFA dataset~\cite{Disfa}. AffectNet is a large-scale faces-in-the-wild dataset manually labeled with eight emotion categories. This dataset has around 288K training and 4K validation images. DISFA is a widely-used spontaneous facial actions dataset manually labeled with the presence/absence of 12 action units~\footnote{The frames with action unit intensities greater than 2 are treated as positives and the remaining are treated as negatives.}. This dataset has around 260K images, out of which 212K images are used for training and 48K images are used for validation. We create four expression embeddings using these two classification networks:
\begin{itemize}[topsep=5pt]
\itemsep 0pt
    \item \textit{AFFNet-CL-P} and \textit{AFFNet-CL-F}: Penultimate and final layer outputs of AFFNet-CL. 
    \item \textit{FACSNet-CL-P} and \textit{FACSNet-CL-F}: Penultimate and final layer outputs of FACSNet-CL.
\end{itemize}

Another way to learn an embedding using a classification dataset is to train an embedding network with a category label-based triplet loss similar to~\cite{FaceNet}. So, we also train an embedding network (referred to as \textit{AFFNet-TL}) on AffectNet dataset using triplet loss. 

For a fair comparison, the input and architecture of all the networks are chosen to be same as FECNet (Figure~\ref{fig:network}) except that the embedding layer is replaced by a softmax classifier for AFFNet-CL and separate binary classifiers for FACSNet-CL.

\subsection{Training and validation}
We define \textit{triplet prediction accuracy} as the percentage of triplets for which the distance (in the embedding space) between the visually most similar pair is less than the distances of these two images from the third. As for the validation measure during training, we use triplet prediction accuracy on the FEC test set for FECNet, (following~\cite{Fabnet}) average area under ROC curve (AUC-ROC) on the AffectNet validation set for AFFNet-CL and AFFNet-TL~\footnote{Nearest neighbor classifier with 800 neighbors is used.}, and (following~\cite{FAUECCV18,AUDandAlign}) average F1-score on the DISFA validation set for FACSNet-CL.

For all the networks, the parameters of the FaceNet layers were kept fixed and the newly-added DenseNet and FC layers were trained starting from Xavier initialization~\cite{Xavier} using Adam optimizer~\cite{Adam} with a learning rate of $5\times10^{-4}$ and dropout of 0.5. FECNet was trained on the FEC training set with mini-batches of 90 samples (30 triplets from each of the triplet types) for 50K iterations, AFFNet-CL and AFFNet-TL were trained on the AffectNet training set with mini-batches of 128 samples (16 samples from each of the eight emotion categories) for 10K iterations, and FACSNet-CL was trained on the DISFA training set with mini-batches of 130 samples (at least 10 positive samples for each action unit and 10 samples with no action units) for 20K iterations. For training FECNet, the value of margin $\delta$ was set to 0.1 for one-class triplets, and 0.2 for two-class and three-class triplets. For training AFFNet-TL, the loss margin was set to 0.2 and the embedding dimensionality was set to 16. All the hyper-parameter values were chosen based on the corresponding validation measures. 

\subsection{Average human performance}
\begin{figure}
    \centering
    \includegraphics[trim={15px 0px 0px 20px},clip,width=0.48\textwidth,height=0.25\textwidth]{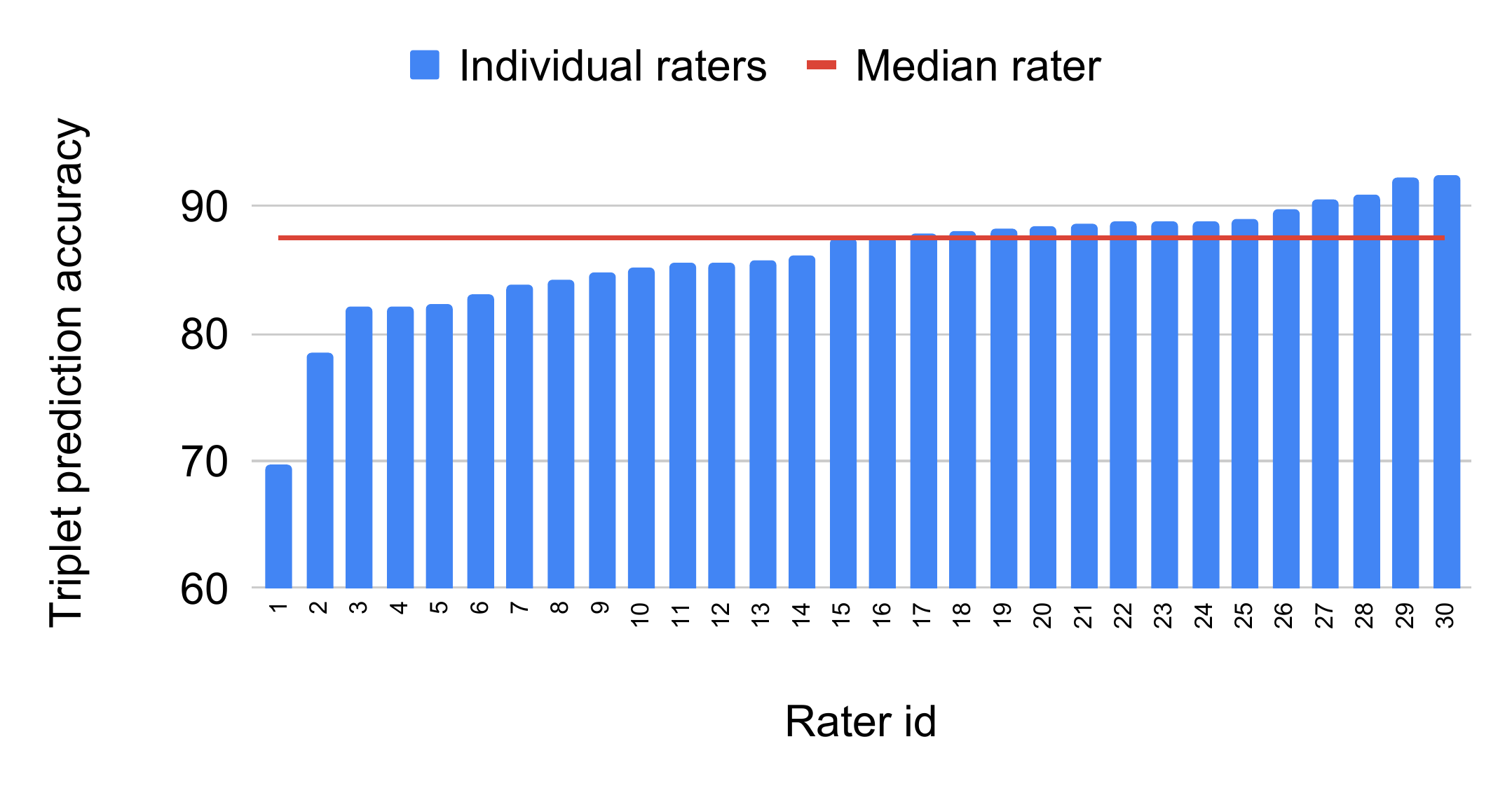}
    \caption{Triplet prediction accuracy of individual raters on the FEC test set (blue bars). The red line shows the median accuracy.}
    \label{fig:human-accuracy}
\end{figure}

To estimate how good humans are at identifying the most similar pair in a triplet, we computed the triplet prediction accuracy values for individual raters based on how often they agree with the maximum-voted label. Figure~\ref{fig:human-accuracy} shows the accuracy values for all the 30 raters who contributed to the FEC test set annotations. The mean and median values are 86.2\% and 87.5\%.

\begin{figure}
    \centering
    \includegraphics[trim={30px 0px 20px 0px},clip,scale=0.32]{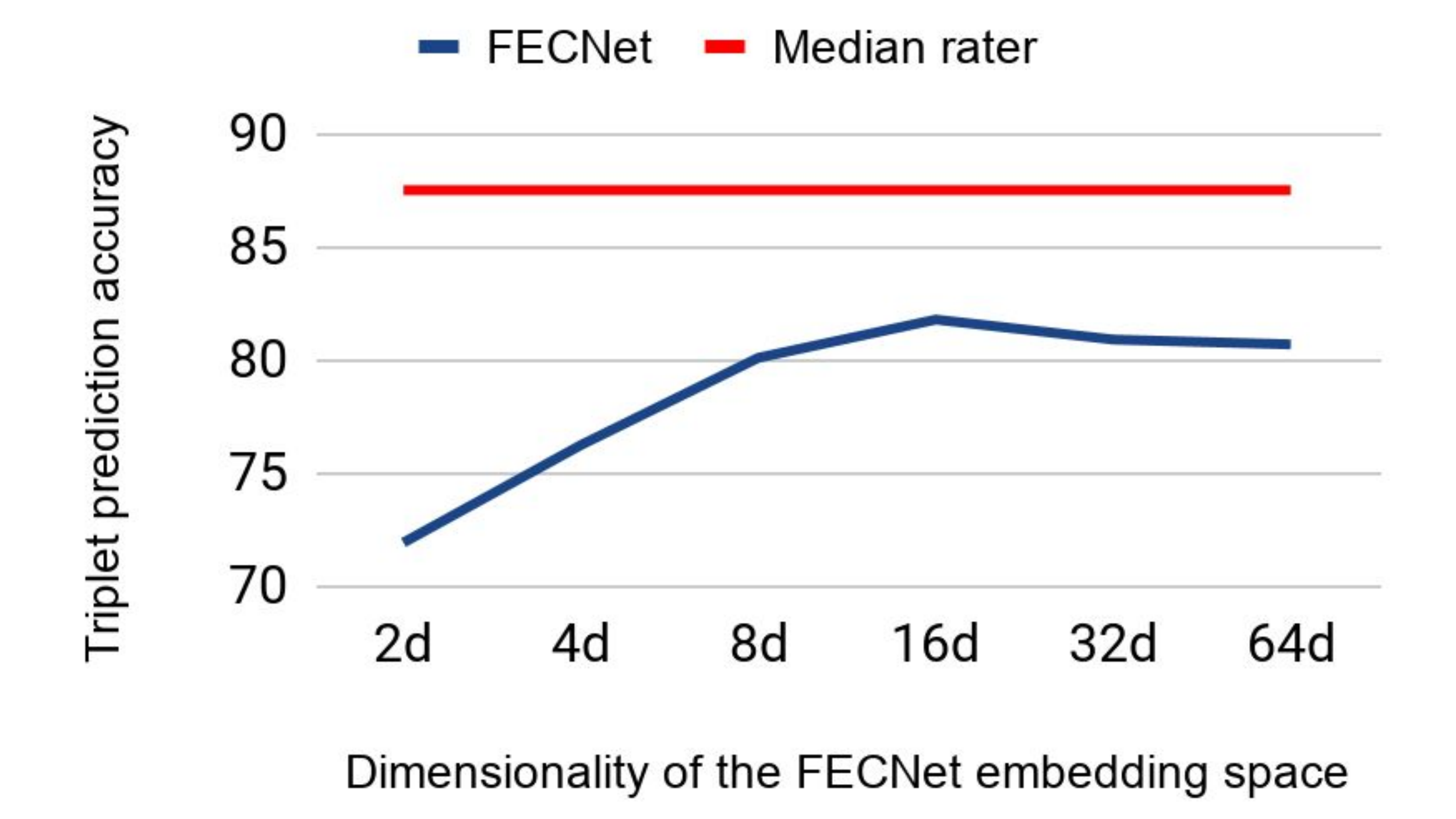}
    \caption{Triplet prediction accuracy on the FEC test set as a function of the FECNet embedding dimensionality (blue curve). The red line corresponds to median rater.}
    \label{fig:dimensions}
\end{figure}
\subsection{Dimensionality of the FECNet embedding}
While we want to represent facial expressions in a continuous fashion using an embedding, it is unclear how many dimensions should be used for the embedding space. To answer this question, we trained FECNet for different values of the output dimensionality. Figure~\ref{fig:dimensions} shows how the triplet prediction accuracy on the FEC test set varies with the dimensionality of the embedding space. The accuracy increases till 16 dimensions and drops slightly after that. Based on these results, we choose 16-dimensions to represent the expression embedding space (referred to as FECNet-16d).

Figure~\ref{fig:dimensions} also shows the median rater accuracy. Using 16 dimensions, the proposed FECNet is able to achieve an accuracy of 81.8\%, which is fairly close to the median rater accuracy (87.5\%).  Note that the triplet prediction accuracy of random choice is 33.3\%.

\begin{table}[t]
    \centering
    \small
    \begin{tabular}{|c|c|c|c|}
    \hline
        \multirow{2}{*}{Embedding} & \multicolumn{3}{c|}{Distance} \\\cline{2-4}
        & $\ell_1$ & $\ell_2$ & Cosine\\\hline
        FACSNet-CL-F & 47.1 & 47.1 & 40.7\\\hline 
        FACSNet-CL-P & 45.3 & 44.2 & 48.3\\\hline
        AFFNet-CL-F  & 49.0 & 47.7 & 49.0\\\hline 
        AFFNet-CL-P & 52.4 & 51.6 & 53.3\\\hline
        AFFNet-TL & - & 49.6 & -\\\hline
        FECNet-16d & - & 81.8 & - \\\hline
    \end{tabular}
    \caption{Triplet prediction accuracy on the FEC test set.}
    \label{tab:affnet-embeddings}
\end{table}
\subsection{Comparison of different embeddings}
\label{sec::comparison}
Table~\ref{tab:affnet-embeddings} shows the triplet prediction accuracy of various embeddings on the FEC test set using different distance functions. Among all the AFFNet and FACSNet embeddings, the combination of AFFNet-CL-P and cosine distance gives the best accuracy, and hence, we use this combination for comparison with FECNet-16d in the rest of the experiments. It is worth noting that the proposed FECNet-16d (81.8\%) performs significantly better than the best competing approach (AFFNet-CL-P + Cosine distance; 53.3\%). 

We also trained FECNet-16d by adding the triplets with weak rater agreement to the training set, but the test accuracy dropped from 81.8\% to 80.5\%.

\begin{table}[t]
    \centering
    \small
    \begin{tabular}{|c|c|c|c|}
    \hline
        Triplet type & AFFNet-CL-P & FECNet-16d & Median rater \\\hline
        One-class & 49.2 & 77.1 & 85.3\\\hline
        Two-class & 59.8 & 85.1 & 89.3\\\hline
        Three-class & 50.4 & 82.6 & 87.2\\\hline\hline
        All triplets & 53.3 & 81.8 & 87.5\\\hline
    \end{tabular}
    \caption{Triplet prediction accuracy for different types of triplets in the FEC test set.}
    \label{fig:triplet-type-performance}
\end{table}
\subsection{Performance for different triplet types}
Table~\ref{fig:triplet-type-performance} shows the triplet prediction accuracy of median rater, FECNet-16d and AFFNet-CL-P for each triplet type in the FEC test set. As expected, the performance is best (85.1\%) for two-class triplets, which are relatively the easiest ones, and is lowest (77.1\%) for one-class triplets, which are relatively the most difficult ones. 
\subsection{Visualization of the FECNet embedding space}
\begin{figure}
    \centering
    \includegraphics[scale=0.45]{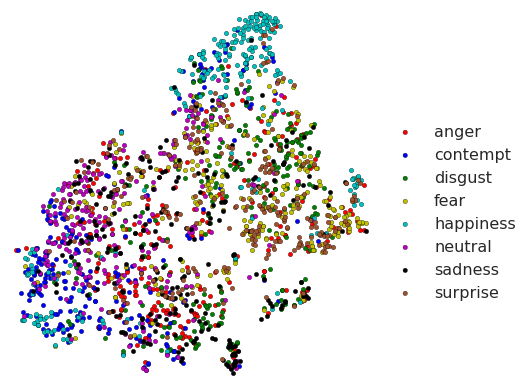}
    \caption{2D visualization of the FECNet-16d embedding space using t-SNE~\cite{tSNE}.}
    \label{fig:tSNE}
\end{figure}
Figure~\ref{fig:tSNE} shows a 2D t-SNE~\cite{tSNE} visualization of the learned FECNet-16d embedding space using the AffectNet validation set. The amount of overlap between two categories in this figure roughly tells us about the extent of visual similarity between them. For example, fear and surprise have a high overlap indicating that they could be confused easily, and both of them have a very low overlap with contempt indicating that they are visually very distinct from contempt. Also, the spread of a category in this figure tells us about the visual diversity within that category. For example, happiness category maps to three distinct regions indicating that there are three visually distinct modes within this category. See Figure~\ref{fig:tsne_faces} for a visualization of the face images that fall into different regions in Figure~\ref{fig:tSNE}.
\begin{table*}[t]
    \centering
    \begin{tabular}{|c|c|c|c|c|c|c|c|c|c|c|}
    \hline
        Album & BO & CB & DT & GB & HC & JL & JC & KM & LJ & LS \\\hline
        FECNet-16d vs AFFNet-CL-P & 5-2 & 9-1 & 5-1 & 9-0 & 10-0 & 9-0 & 7-1 & 10-0 & 1-4 & 1-6 \\\hline
    \end{tabular}
    \caption{Number of votes received by the summaries generated by the FECNet-16d and AFFNet-CL-P embeddings.}
    \label{tab:clustering_results}
\end{table*}
\subsection{Applications}
\subsubsection{Image retrieval}
We can perform expression-based image retrieval by using nearest neighbor search in the expression embedding space. To compare the retrieval performance of FECNet-16d and AFFNet-CL-P embeddings, we use a query set consisting of 25 face images and a database (CelebA dataset~\cite{CelebA}) consisting of 200K face images. For each query, we retrieved $N$ nearest neighbors ($N$ varied from 1 to 10) using both the embeddings and ranked the $2N$ retrieved images based on how close they are to the query as judged by ten human raters. Since ranking all $2N$ images at once is difficult for human raters, we asked them to rank two images at a time. In each pairwise ranking, the winner and looser get a score of +1 and -1, respectively. If it is a tie, i.e., the two images get equal number of rater votes, then both of them get a score of zero. We obtained such pairwise ranking scores for all pairs and converted them into a global ranking based on the overall scores. 

\begin{figure}[t]
    \centering
    \includegraphics[trim={100px 10px 50px 10px},scale=0.35]{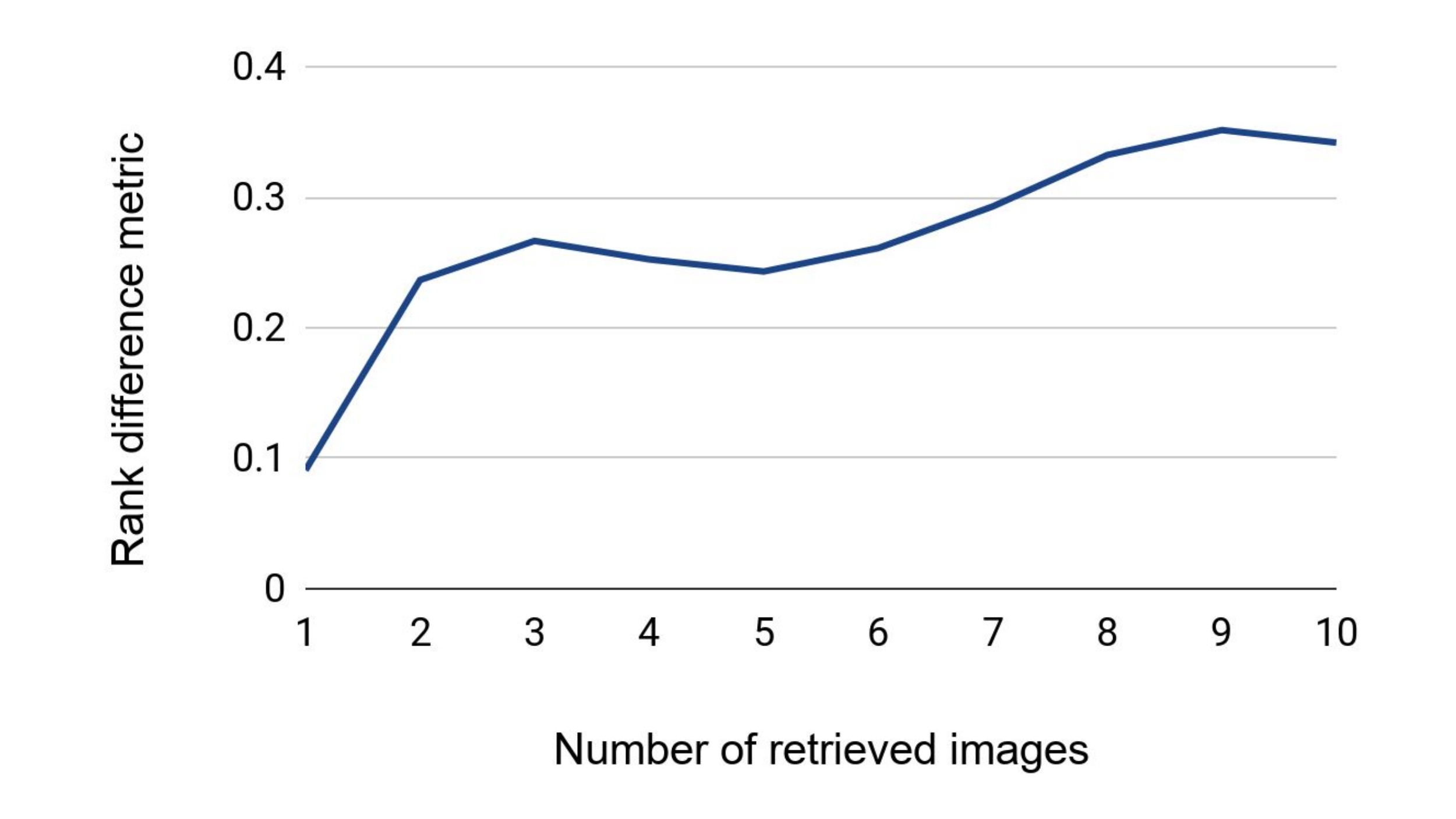}
    \caption{Rank-difference metric as a function of the number of retrieved images $N$.}
    \label{fig:rank-difference}
\end{figure}

For numerical evaluation, we use \textit{rank-difference} metric, defined as the average difference in the ranks of images retrieved by AFFNet-CL-P and FECNet-16d embeddings, respectively, divided by the number of retrieved images $N$. Positive value of this rank-difference metric indicates that FECNet-16d embedding is better than AFFNet-CL-P embedding. The lowest value for this metric is $-1$, corresponding to the case when all the AFFNet-CL-P retrieval results are ranked lower than all the FECNet-16d retrieval results, and the highest value is $+1$, corresponding to the case when all the FECNet-16d retrieval results are ranked lower than all the AFFNet-CL-P retrieval results. Figure~\ref{fig:rank-difference} shows the rank-difference metric for different values of $N$. Positive value of the metric for all values of $N$ clearly indicates that the proposed FECNet-16d embedding produces better matches compared to the AFFNet-CL-P embedding.

Figure~\ref{fig:retrieval_results} shows the top-5 retrieved images for some of the queries. The overall results of the proposed FECNet-16d embedding are clearly better than the results of AFFNet-CL-P embedding. Specifically, the FECNet-16d embedding pays attention to finer details such as teeth-not-visible (first query), eyes-closed (second and third queries) and looking straight (fourth query). See Figures~\ref{fig:retrieval-first} to~\ref{fig:retrieval-last} for additional retrieval results.

\begin{figure}[t]
    \centering
    \includegraphics[trim={20px 10px 50px 20px},clip,scale=0.37]{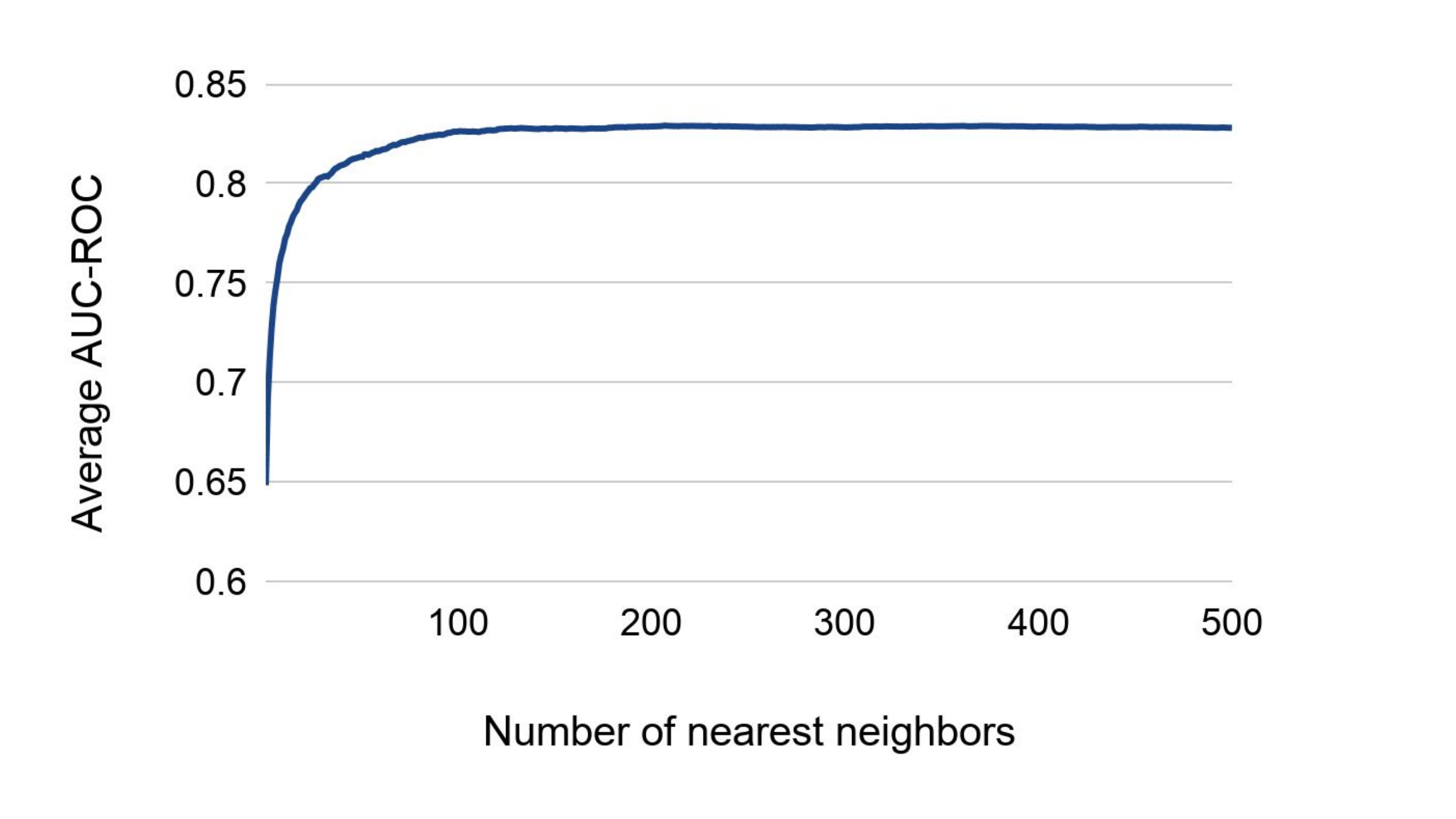}
    \caption{Classification performance of the FECNet-16d embedding on the AffectNet validation set when combined with K-NN classifier.}
    \label{fig:classification}
\end{figure}
\subsubsection{Photo album summarization}
In this task, we are interested in summarizing the diverse expression content present in a given photo album using a fixed number of images. Expression embedding can be used for this task by combining it with a clustering algorithm.

For evaluation, we created ten photo albums (100-200 images in each album) by downloading images of ten celebrities using Google image search. For each album, we ran hierarchical agglomerative clustering~\footnote{Cosine distance and maximum linkage were used.} (10 clusters) with FECNet-16d and AFFNet-CL-P embeddings, and used the images that are closest to cluster centers for generating the summaries. We showed these two summaries to ten human raters and asked them which one is better. The raters were also allowed to choose the \textit{difficult-to-decide} option. Table~\ref{tab:clustering_results} shows the number of votes received by both the embeddings for all the albums. Humans prefer the summaries generated by the proposed FECNet-16d embedding for eight out of ten albums. Figures~\ref{fig:summary1} and~\ref{fig:summary2} show the summary results for all the albums. We can see that the expression content is more diverse in the summaries produced by the FECNet-16d embedding for most of the albums. 
\begin{table*}[t]
\centering
\small
\begin{tabular}{|c|c|c|c|c|c|c|c|c|c|}
    \hline
        Approach & Neutral & Happiness & Sadness & Surprise & Fear & Disgust & Anger & Contempt & Average\\\hline
        AFFNet-CL & 84.6 & \textbf{96.5} & \textbf{90.7} & \textbf{88.5} & \textbf{90.2} & \textbf{85.2} & \textbf{88.3} & \textbf{85.0} & \textbf{88.6}\\\hline
        AFFNet-TL & \textbf{85.9} & 96.0 & 89.2 & \textbf{88.5} & 89.6 & 83.7 & 87.9 & 82.6 & 87.9\\\hline
        FECNet-16d + K-NN & 83.3 & 94.9 & 78.0 & 83.0 & 84.5 & 79.3 & 78.7 & 81.2 & 82.9\\\hline
        AlexNet~\cite{AffectNet} & - & - & - & - & - & - & - & - & 82.0\\\hline
        VGG-Face descriptor~\cite{Fabnet} & 75.9 & 92.2 & 80.5 & 81.4 & 82.3 & 81.4 & 81.2 & 77.1 & 81.5\\\hline
        FAb-Net~\cite{Fabnet} & 72.3 & 90.4 & 70.9 & 78.6 & 77.8 & 72.5 & 76.4 & 72.2 & 76.4\\\hline
\end{tabular}
\caption{Expression classification results (AUC-ROC) on the AffectNet~\cite{AffectNet} validation set.}
\label{tab:affnet-classification}
\vspace{10pt}
\includegraphics[width=\textwidth, height=0.5\textwidth]{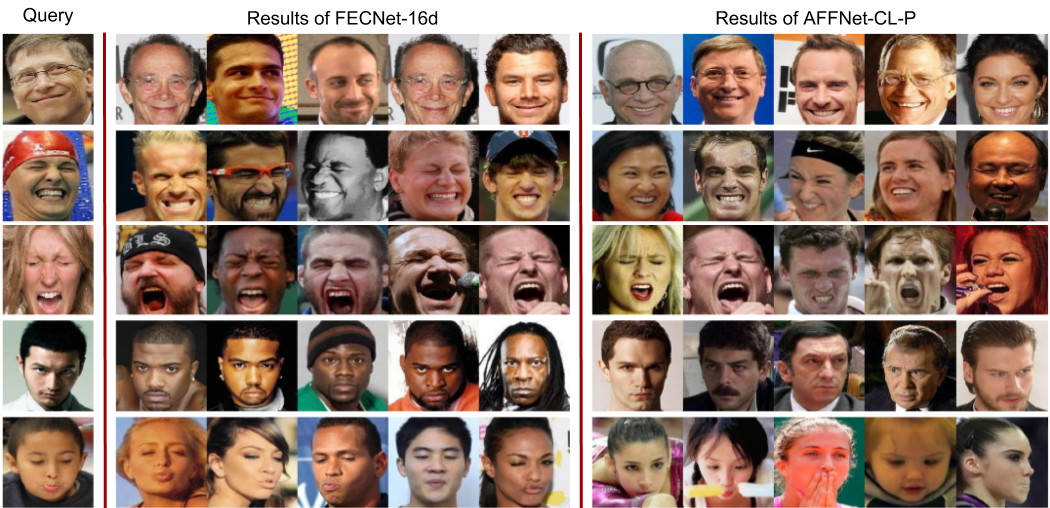}
\captionof{figure}{Top-5 images retrieved using FECNet-16d (left) and AFFNet-CL-P (right) embeddings. The overall results of FECNet-16d match the query set better when compared to AFFNet-CL-P.}
\label{fig:retrieval_results}
\end{table*}
\subsubsection{Emotion classification}
The proposed FECNet-16d embedding can be used for emotion classification by combining it with K-Nearest Neighbor (K-NN) classifier. Figure~\ref{fig:classification} shows the average AUC-ROC of the FECNet-16d embedding on the AffectNet validation set as a function of the number of neighbors used. The performance increases up to 200 neighbors and then remains stable. Table~\ref{tab:affnet-classification} compares the classification performance of the FECNet-16d embedding (using 200 neighbors) with other approaches. Note that AFFNet-CL and AFFNet-TL have the same architecture as FECNet-16d and are specifically trained for classification using AffectNet training data. Hence, as expected, they perform a bit better than FECNet-16d. However, despite not being trained for classification, the FECNet-16d embedding outperforms AlexNet and VGG-Face based classifiers, demonstrating that it is well-suited for classification.

\section{Conclusions and Future Work}
In this work, we presented the first large-scale facial expression comparison dataset annotated by human raters, and learned a compact (16-dimensional) facial expression embedding using this dataset with triplet loss. The embedding learned using this dataset performs better than various other embeddings learned using existing emotion and action units datasets. We experimentally demonstrated the usefulness of the proposed embedding for various applications such as expression retrieval, photo album summarization, and emotion classification.

Another interesting application of the FECNet embedding is hard-negative mining for expression classification. Since FECNet is trained using human visual preferences, negative samples that are close to the positive samples in the FECNet embedding space can be considered as hard negatives while training a classification model. We plan to explore this further in our future work.
\section*{Acknowledgements}
We thank Gautam Prasad, Ting Liu, Brendan Jou, Alan Cowen, Florian Schroff and Hartwig Adam from Google for their support and suggestions during the data collection process.
\clearpage
\begin{minipage}{\textwidth}
    \centering
    \includegraphics[scale=0.13]{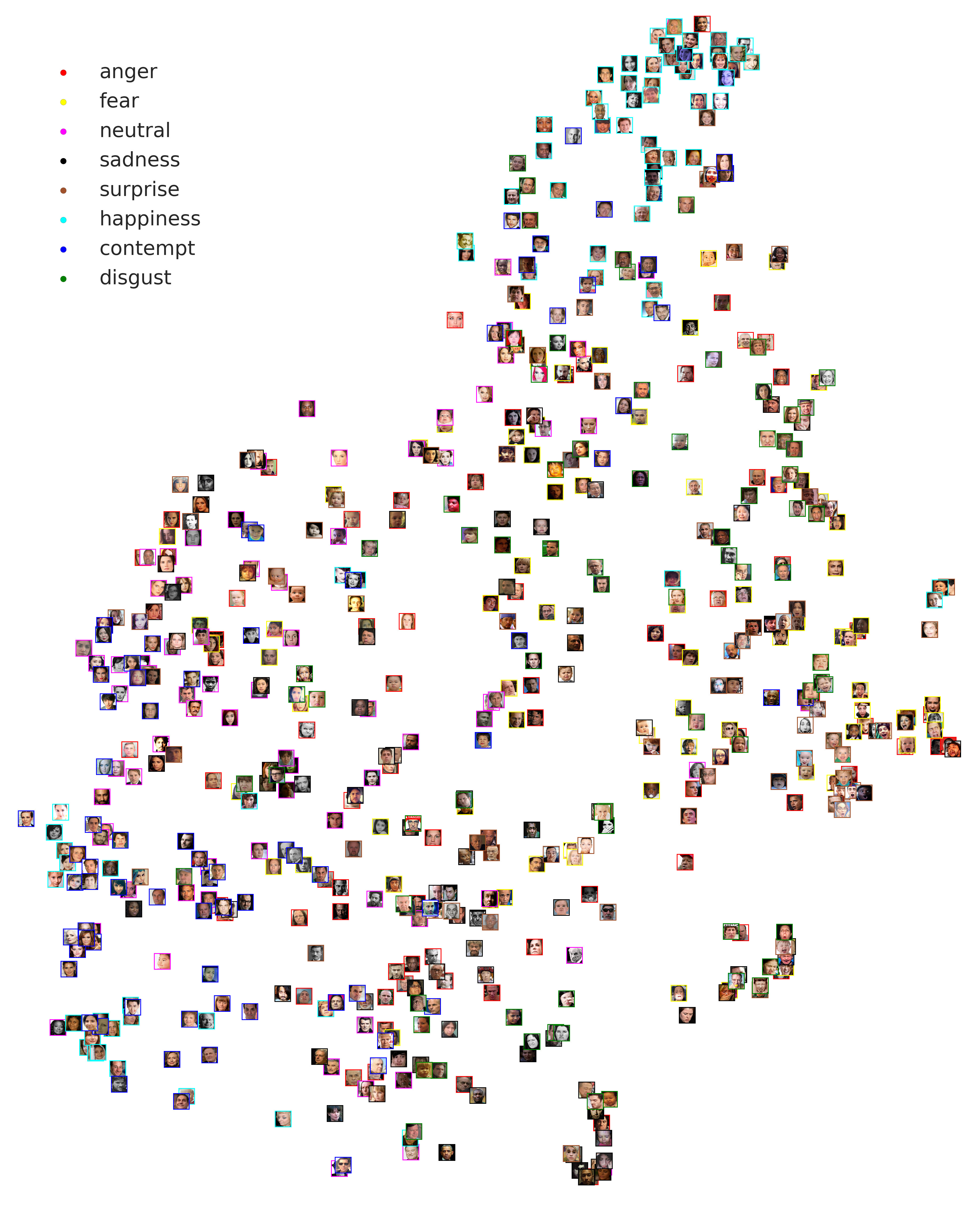}
    \captionof{figure}{2D visualization of the learned FECNet-16d embedding space using t-SNE~\cite{tSNE}. The color of the bounding box of a face represents the emotion label of that face. Please zoom in to see the face images clearly.}
    \label{fig:tsne_faces}
\end{minipage}

\begin{table*}
\begin{minipage}[t]{\textwidth}
\centering
\begin{minipage}[t]{0.49\textwidth}
\centering
\includegraphics[width=8cm, height=3.5cm]{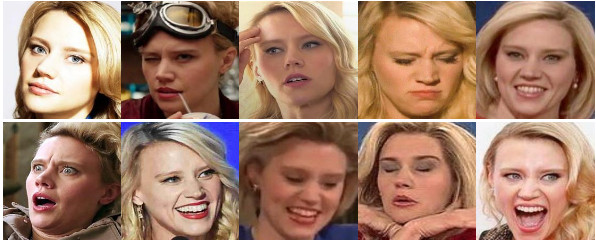}
\end{minipage}
\begin{minipage}[t]{0.49\textwidth}
\centering
\includegraphics[width=8cm, height=3.5cm]{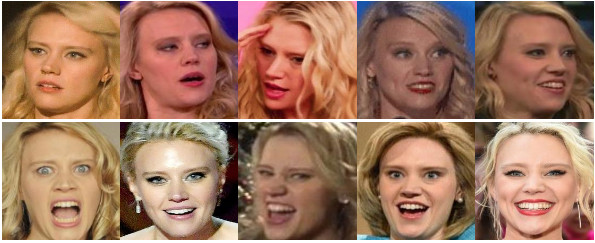}
\end{minipage}
\caption*{Kate McKinnon (KM)}
\vspace{10pt}
\end{minipage}

\begin{minipage}[t]{\textwidth}
\centering
\begin{minipage}[t]{0.49\textwidth}
\centering
\includegraphics[width=8cm, height=3.5cm]{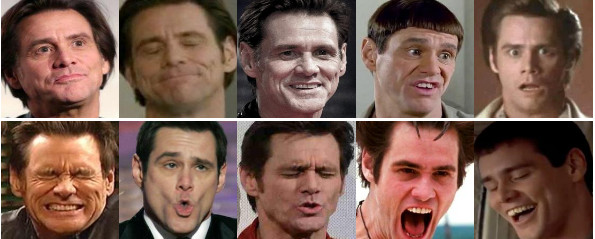}
\end{minipage}
\begin{minipage}[t]{0.49\textwidth}
\centering
\includegraphics[width=8cm, height=3.5cm]{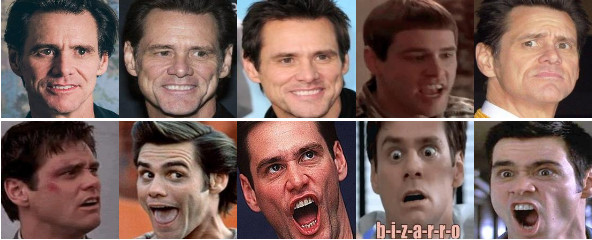}
\end{minipage}
\caption*{Jim Carrey (JC)}
\vspace{10pt}
\end{minipage}

\begin{minipage}[t]{\textwidth}
\centering
\begin{minipage}[t]{0.49\textwidth}
\centering
\includegraphics[width=8cm, height=3.5cm]{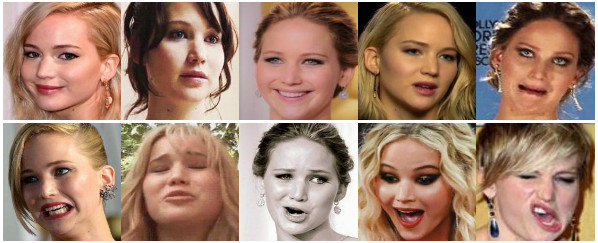}
\end{minipage}
\begin{minipage}[t]{0.49\textwidth}
\centering
\includegraphics[width=8cm, height=3.5cm]{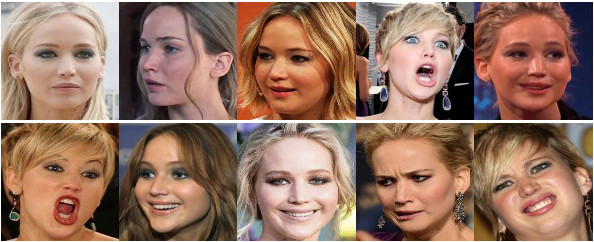}
\end{minipage}
\caption*{Jennifer Lawrence (JL)}
\vspace{10pt}
\end{minipage}

\begin{minipage}[t]{\textwidth}
\centering
\begin{minipage}[t]{0.49\textwidth}
\centering
\includegraphics[width=8cm, height=3.5cm]{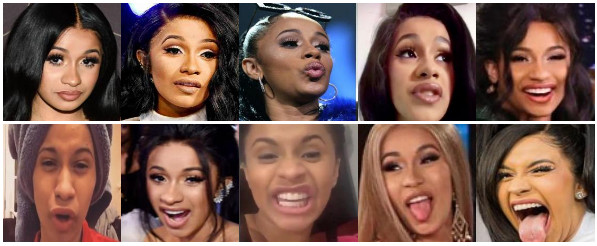}
\end{minipage}
\begin{minipage}[t]{0.49\textwidth}
\centering
\includegraphics[width=8cm, height=3.5cm]{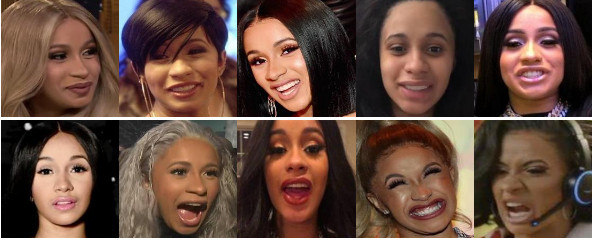}
\end{minipage}
\caption*{Cardi B (CB)}
\vspace{10pt}
\end{minipage}

\begin{minipage}[t]{\textwidth}
\centering
\begin{minipage}[t]{0.49\textwidth}
\centering
\includegraphics[width=8cm, height=3.5cm]{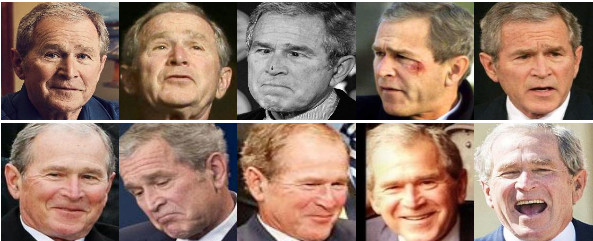}
\end{minipage}
\begin{minipage}[t]{0.49\textwidth}
\centering
\includegraphics[width=8cm, height=3.5cm]{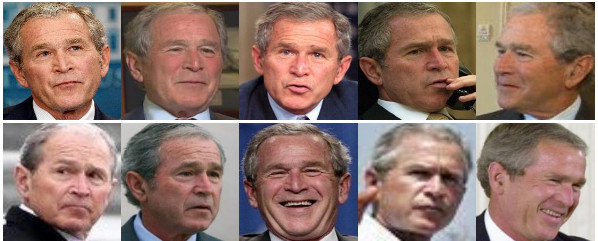}
\end{minipage}
\caption*{George Bush (GB)}
\vspace{10pt}
\end{minipage}
\captionof{figure}{Summaries generated by the FECNet-16d (left) and AFFNet-CL-P (right) embeddings.}
\label{fig:summary1}
\end{table*}

\begin{table*}
\begin{minipage}[t]{\textwidth}
\centering
\begin{minipage}[t]{0.49\textwidth}
\centering
\includegraphics[width=8cm, height=3.5cm]{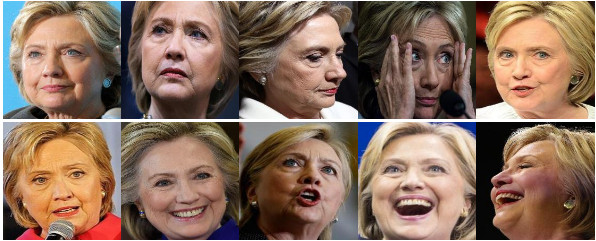}
\end{minipage}
\begin{minipage}[t]{0.49\textwidth}
\centering
\includegraphics[width=8cm, height=3.5cm]{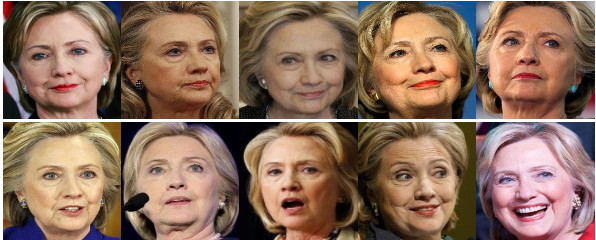}
\end{minipage}
\caption*{Hillary Clinton (HC)}
\vspace{10pt}
\end{minipage}

\begin{minipage}[t]{\textwidth}
\centering
\begin{minipage}[t]{0.49\textwidth}
\centering
\includegraphics[width=8cm, height=3.5cm]{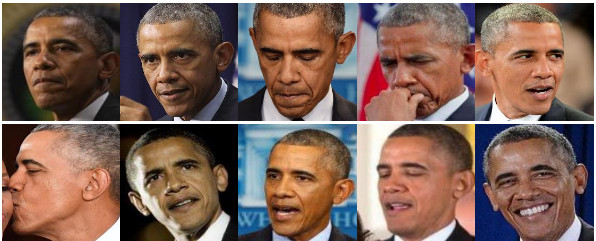}
\end{minipage}
\begin{minipage}[t]{0.49\textwidth}
\centering
\includegraphics[width=8cm, height=3.5cm]{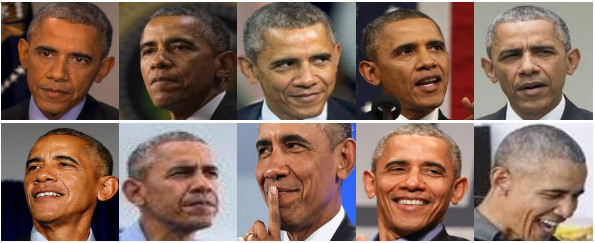}
\end{minipage}
\caption*{Barack Obama (BO)}
\vspace{10pt}
\end{minipage}

\begin{minipage}[t]{\textwidth}
\centering
\begin{minipage}[t]{0.49\textwidth}
\centering
\includegraphics[width=8cm, height=3.5cm]{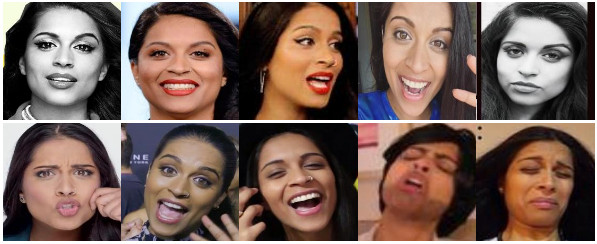}
\end{minipage}
\begin{minipage}[t]{0.49\textwidth}
\centering
\includegraphics[width=8cm, height=3.5cm]{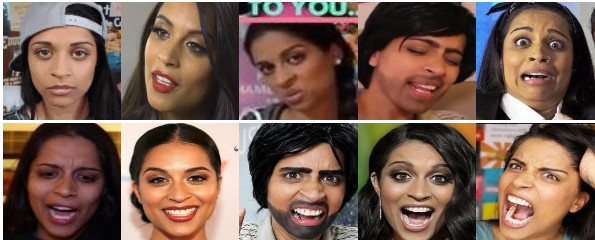}
\end{minipage}
\caption*{Lilly Singh (LS)}
\vspace{10pt}
\end{minipage}

\begin{minipage}[t]{\textwidth}
\centering
\begin{minipage}[t]{0.49\textwidth}
\centering
\includegraphics[width=8cm, height=3.5cm]{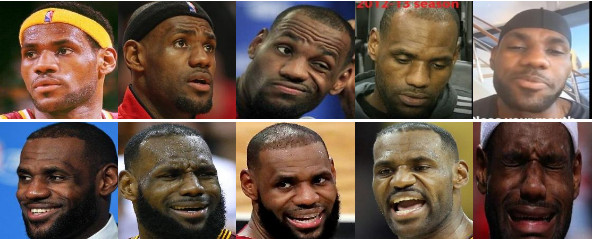}
\end{minipage}
\begin{minipage}[t]{0.49\textwidth}
\centering
\includegraphics[width=8cm, height=3.5cm]{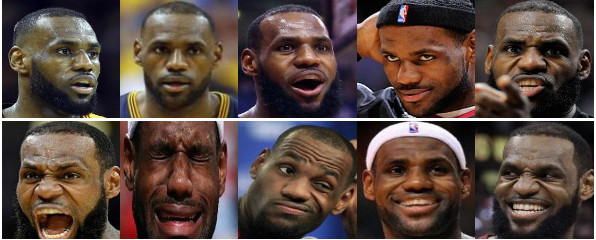}
\end{minipage}
\caption*{Lebron James (LJ)}
\vspace{10pt}
\end{minipage}

\begin{minipage}[t]{\textwidth}
\centering
\begin{minipage}[t]{0.49\textwidth}
\centering
\includegraphics[width=8cm, height=3.5cm]{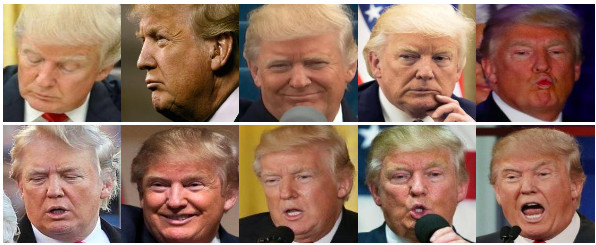}
\end{minipage}
\begin{minipage}[t]{0.49\textwidth}
\centering
\includegraphics[width=8cm, height=3.5cm]{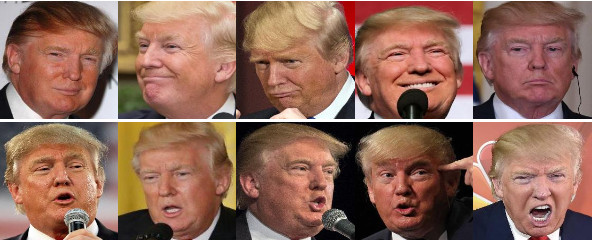}
\end{minipage}
\caption*{Donald Trump (DT)}
\vspace{10pt}
\end{minipage}
\captionof{figure}{Summaries generated by the FECNet-16d (left) and AFFNet-CL-P (right) embeddings.}
\label{fig:summary2}
\end{table*}

\begin{table*}
\begin{minipage}[t]{\textwidth}
\centering
\begin{minipage}[t]{0.1\textwidth}
\centering
\includegraphics[width=0.85\textwidth, height=0.85\textwidth]{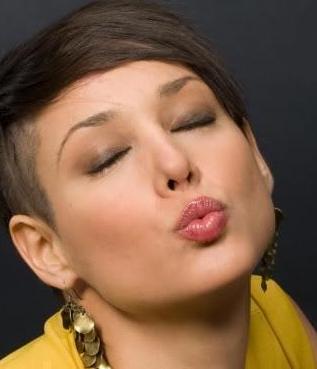}
\end{minipage}
\begin{minipage}[t]{0.89\textwidth}
\centering
\includegraphics[trim={40px 30px 0px 0px}, clip, width=\textwidth]{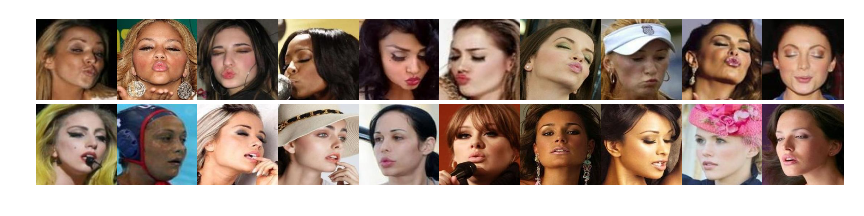}
\end{minipage}
\vspace{15pt}
\end{minipage}

\begin{minipage}[t]{\textwidth}
\centering
\begin{minipage}[t]{0.1\textwidth}
\centering
\includegraphics[width=0.85\textwidth, height=0.85\textwidth]{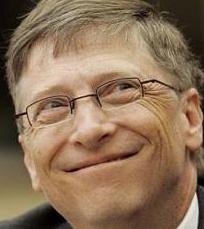}
\end{minipage}
\begin{minipage}[t]{0.89\textwidth}
\centering
\includegraphics[trim={40px 30px 0px 0px}, clip, width=\textwidth]{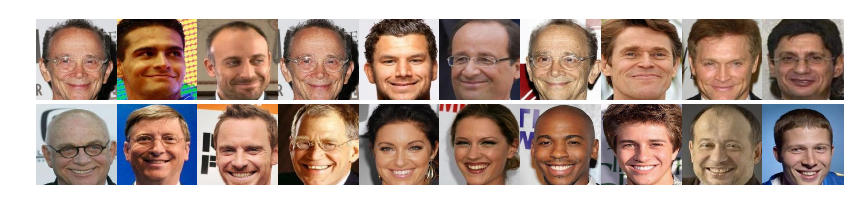}
\end{minipage}
\vspace{15pt}
\end{minipage}

\begin{minipage}[t]{\textwidth}
\centering
\begin{minipage}[t]{0.1\textwidth}
\centering
\includegraphics[width=0.85\textwidth, height=0.85\textwidth]{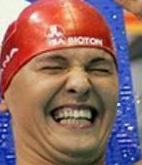}
\end{minipage}
\begin{minipage}[t]{0.89\textwidth}
\centering
\includegraphics[trim={40px 30px 0px 0px}, clip, width=\textwidth]{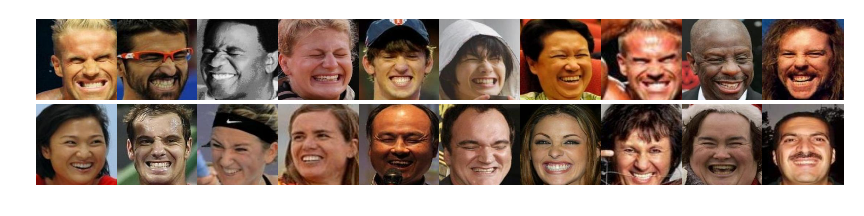}
\end{minipage}
\vspace{15pt}
\end{minipage}

\begin{minipage}[t]{\textwidth}
\centering
\begin{minipage}[t]{0.1\textwidth}
\centering
\includegraphics[width=0.85\textwidth, height=0.85\textwidth]{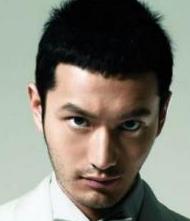}
\end{minipage}
\begin{minipage}[t]{0.89\textwidth}
\centering
\includegraphics[trim={40px 30px 0px 0px}, clip, width=\textwidth]{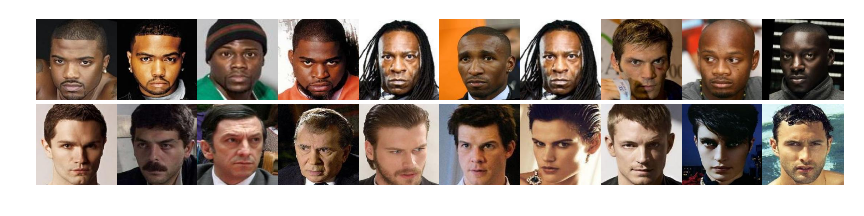}
\end{minipage}
\vspace{15pt}
\end{minipage}

\begin{minipage}[t]{\textwidth}
\centering
\begin{minipage}[t]{0.1\textwidth}
\centering
\includegraphics[width=0.85\textwidth, height=0.85\textwidth]{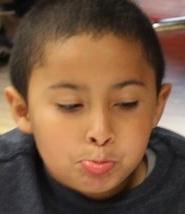}
\end{minipage}
\begin{minipage}[t]{0.89\textwidth}
\centering
\includegraphics[trim={40px 30px 0px 0px}, clip, width=\textwidth]{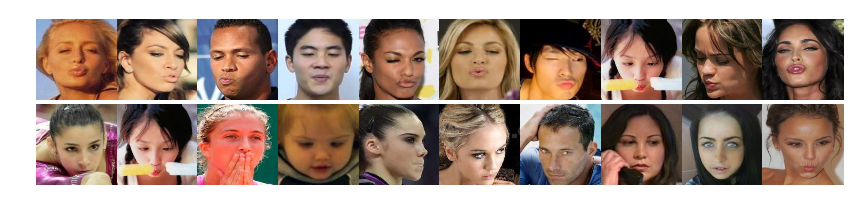}
\end{minipage}
\vspace{15pt}
\end{minipage}
\captionof{figure}{Expression retrieval results - Images on the left are the query expressions. For each query, the top row corresponds to the FECNet-16d embedding and the bottom row corresponds to the
AFFNet-CL-P embedding.}
\label{fig:retrieval-first}
\end{table*}

\begin{table*}
\begin{minipage}[t]{\textwidth}
\centering
\begin{minipage}[t]{0.1\textwidth}
\centering
\includegraphics[width=0.85\textwidth, height=0.85\textwidth]{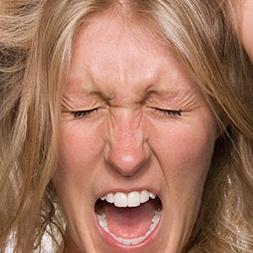}
\end{minipage}
\begin{minipage}[t]{0.89\textwidth}
\centering
\includegraphics[trim={40px 30px 0px 0px}, clip, width=\textwidth]{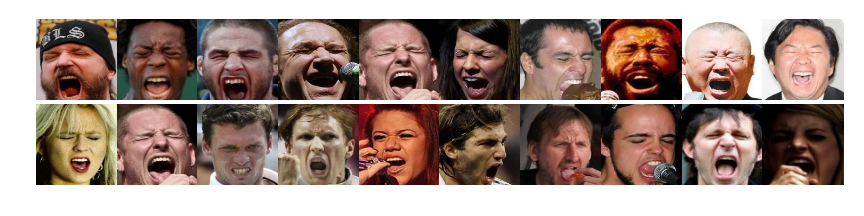}
\end{minipage}
\vspace{15pt}
\end{minipage}

\begin{minipage}[t]{\textwidth}
\centering
\begin{minipage}[t]{0.1\textwidth}
\centering
\includegraphics[width=0.85\textwidth, height=0.85\textwidth]{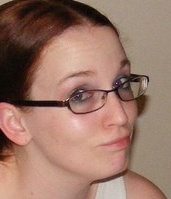}
\end{minipage}
\begin{minipage}[t]{0.89\textwidth}
\centering
\includegraphics[trim={40px 30px 0px 0px}, clip, width=\textwidth]{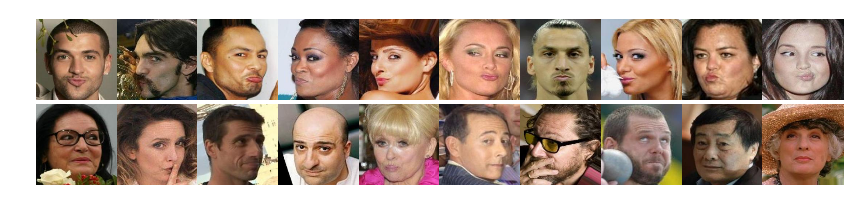}
\end{minipage}
\vspace{15pt}
\end{minipage}

\begin{minipage}[t]{\textwidth}
\centering
\begin{minipage}[t]{0.1\textwidth}
\centering
\includegraphics[width=0.85\textwidth, height=0.85\textwidth]{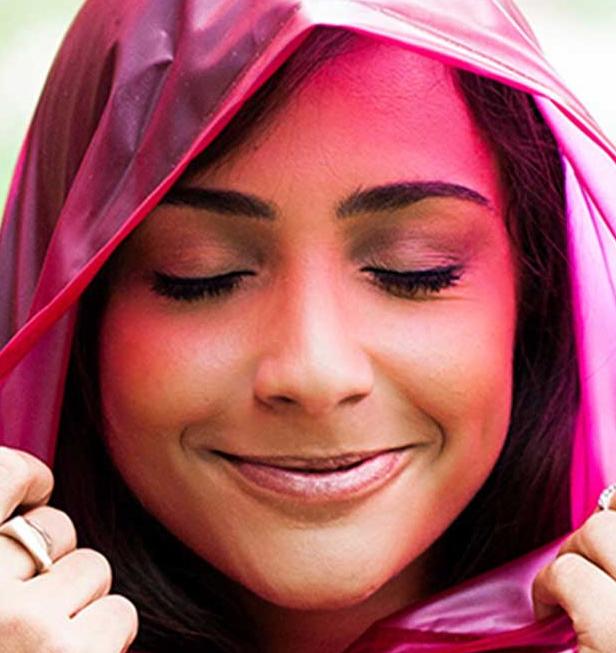}
\end{minipage}
\begin{minipage}[t]{0.89\textwidth}
\centering
\includegraphics[trim={40px 30px 0px 0px}, clip, width=\textwidth]{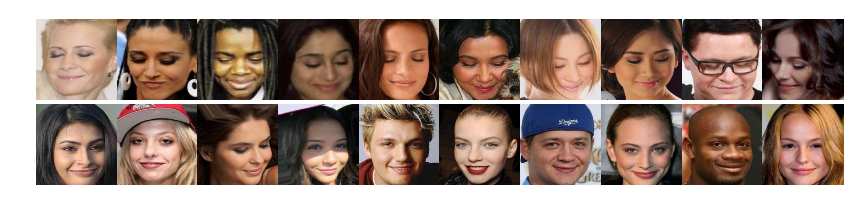}
\end{minipage}
\vspace{15pt}
\end{minipage}

\begin{minipage}[t]{\textwidth}
\centering
\begin{minipage}[t]{0.1\textwidth}
\centering
\includegraphics[width=0.85\textwidth, height=0.85\textwidth]{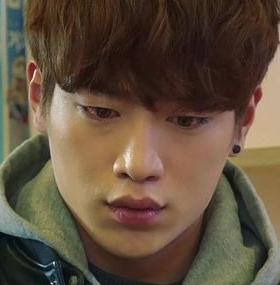}
\end{minipage}
\begin{minipage}[t]{0.89\textwidth}
\centering
\includegraphics[trim={40px 30px 0px 0px}, clip, width=\textwidth]{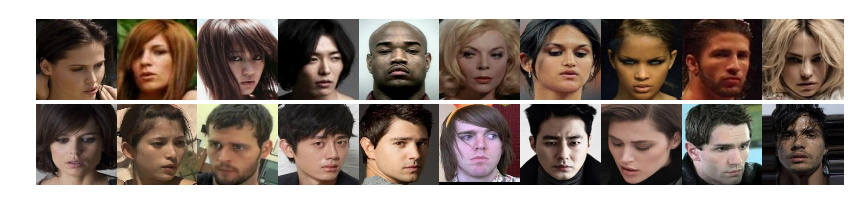}
\end{minipage}
\vspace{15pt}
\end{minipage}

\begin{minipage}[t]{\textwidth}
\centering
\begin{minipage}[t]{0.1\textwidth}
\centering
\includegraphics[width=0.85\textwidth, height=0.85\textwidth]{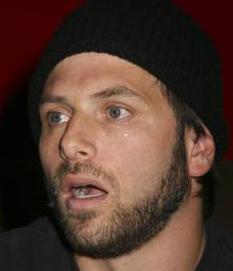}
\end{minipage}
\begin{minipage}[t]{0.89\textwidth}
\centering
\includegraphics[trim={40px 30px 0px 0px}, clip, width=\textwidth]{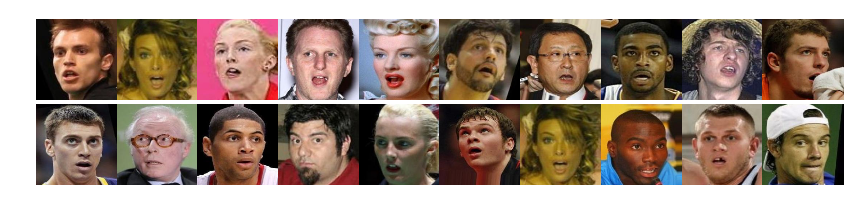}
\end{minipage}
\vspace{15pt}
\end{minipage}

\captionof{figure}{Expression retrieval results - Images on the left are the query expressions. For each query, the top row corresponds to the FECNet-16d embedding and the bottom row corresponds to the
AFFNet-CL-P embedding.}
\end{table*}

\begin{table*}
\begin{minipage}[t]{\textwidth}
\centering
\begin{minipage}[t]{0.1\textwidth}
\centering
\includegraphics[width=0.85\textwidth, height=0.85\textwidth]{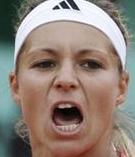}
\end{minipage}
\begin{minipage}[t]{0.89\textwidth}
\centering
\includegraphics[trim={40px 30px 0px 0px}, clip, width=\textwidth]{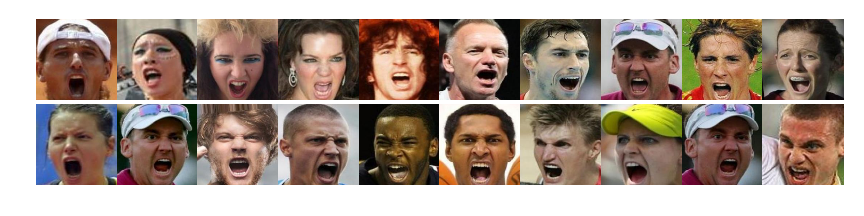}
\end{minipage}
\vspace{15pt}
\end{minipage}

\begin{minipage}[t]{\textwidth}
\centering
\begin{minipage}[t]{0.1\textwidth}
\centering
\includegraphics[width=0.85\textwidth, height=0.85\textwidth]{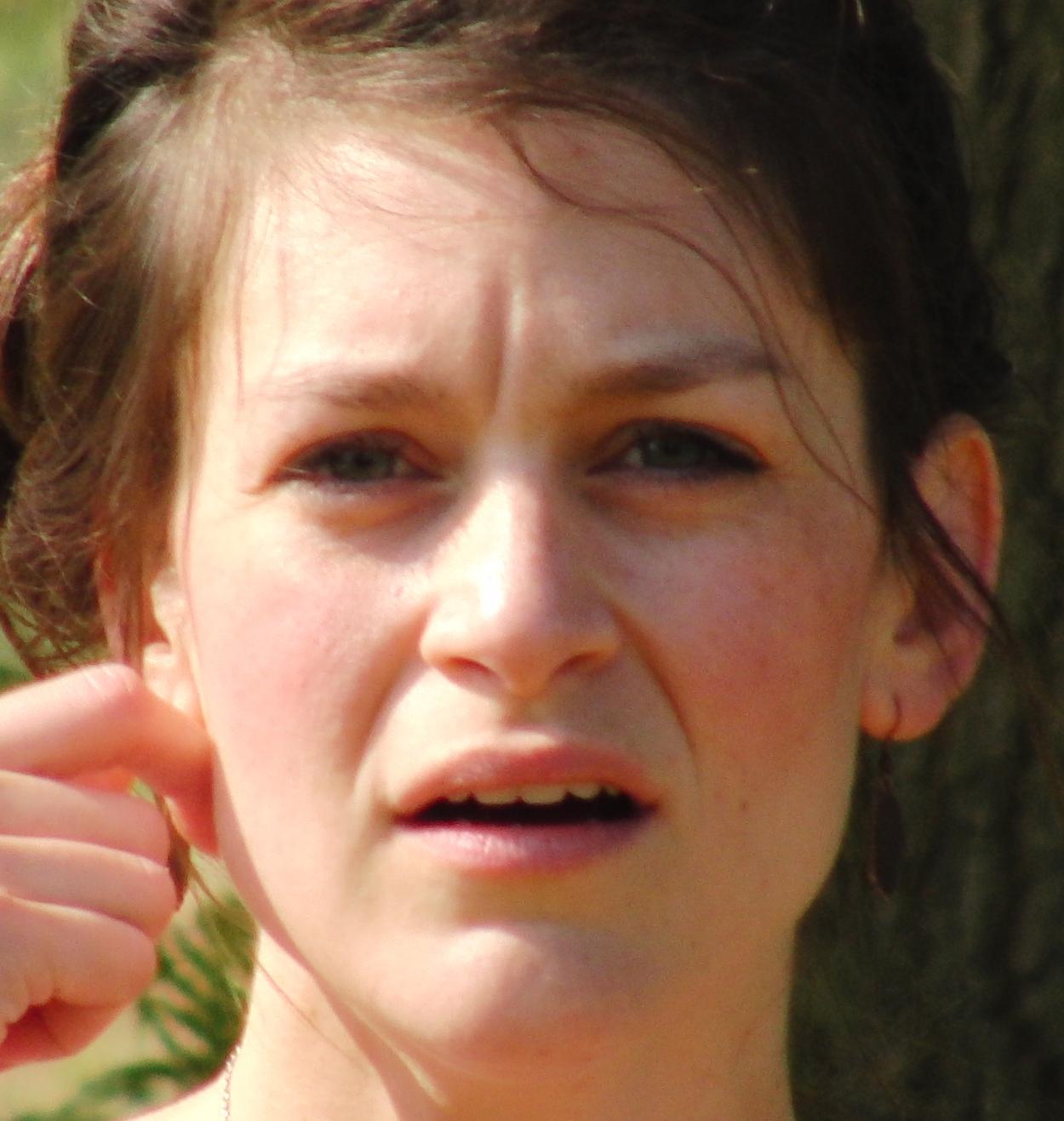}
\end{minipage}
\begin{minipage}[t]{0.89\textwidth}
\centering
\includegraphics[trim={40px 30px 0px 0px}, clip, width=\textwidth]{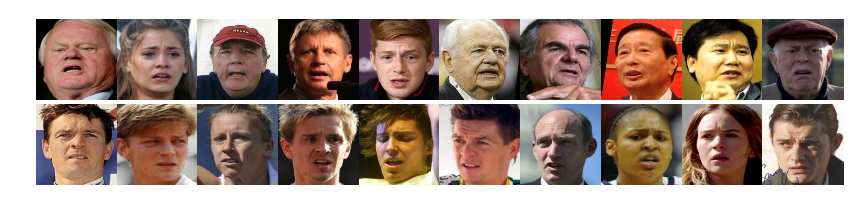}
\end{minipage}
\vspace{15pt}
\end{minipage}

\begin{minipage}[t]{\textwidth}
\centering
\begin{minipage}[t]{0.1\textwidth}
\centering
\includegraphics[width=0.85\textwidth, height=0.85\textwidth]{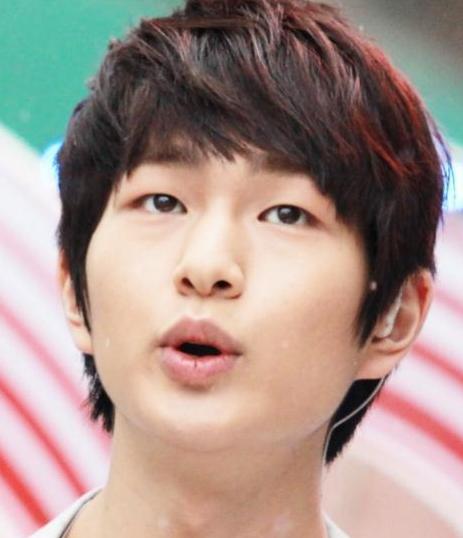}
\end{minipage}
\begin{minipage}[t]{0.89\textwidth}
\centering
\includegraphics[trim={40px 30px 0px 0px}, clip, width=\textwidth]{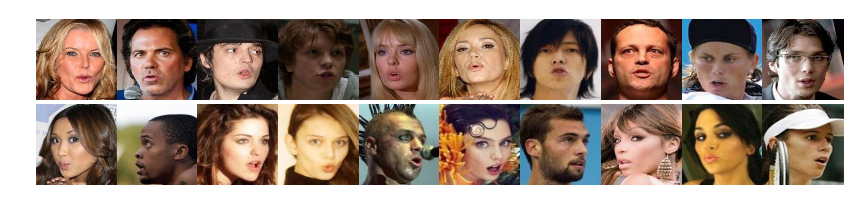}
\end{minipage}
\vspace{15pt}
\end{minipage}

\begin{minipage}[t]{\textwidth}
\centering
\begin{minipage}[t]{0.1\textwidth}
\centering
\includegraphics[width=0.85\textwidth, height=0.85\textwidth]{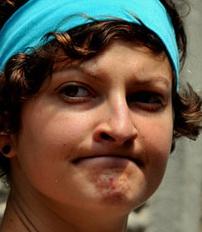}
\end{minipage}
\begin{minipage}[t]{0.89\textwidth}
\centering
\includegraphics[trim={40px 30px 0px 0px}, clip, width=\textwidth]{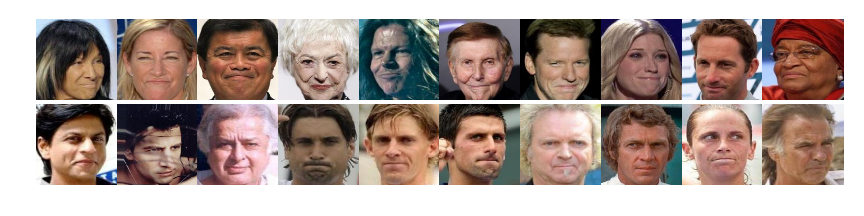}
\end{minipage}
\vspace{15pt}
\end{minipage}

\begin{minipage}[t]{\textwidth}
\centering
\begin{minipage}[t]{0.1\textwidth}
\centering
\includegraphics[width=0.85\textwidth, height=0.85\textwidth]{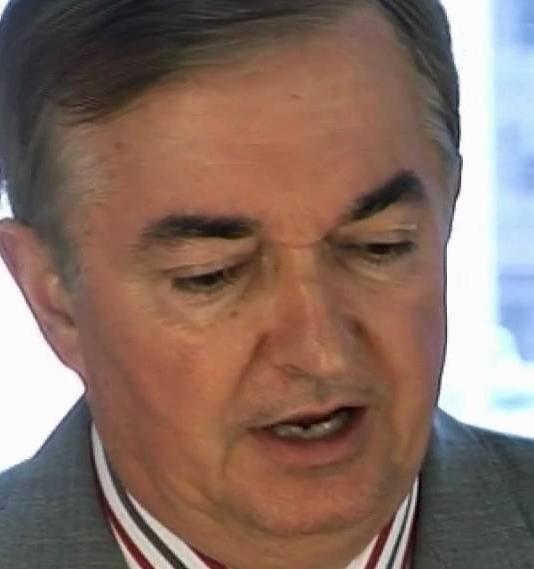}
\end{minipage}
\begin{minipage}[t]{0.89\textwidth}
\centering
\includegraphics[trim={40px 30px 0px 0px}, clip, width=\textwidth]{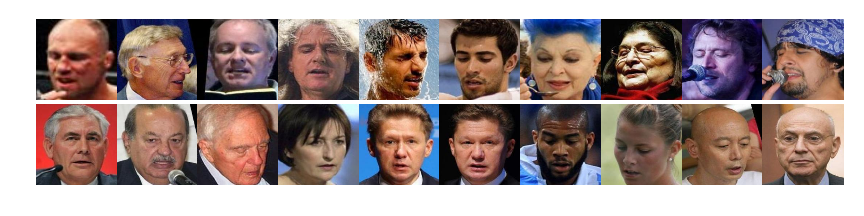}
\end{minipage}
\vspace{15pt}
\end{minipage}

\captionof{figure}{Expression retrieval results - Images on the left are the query expressions. For each query, the top row corresponds to the FECNet-16d embedding and the bottom row corresponds to the
AFFNet-CL-P embedding.}
\end{table*}

\begin{table*}

\begin{minipage}[t]{\textwidth}
\centering
\begin{minipage}[t]{0.1\textwidth}
\centering
\includegraphics[width=0.85\textwidth, height=0.85\textwidth]{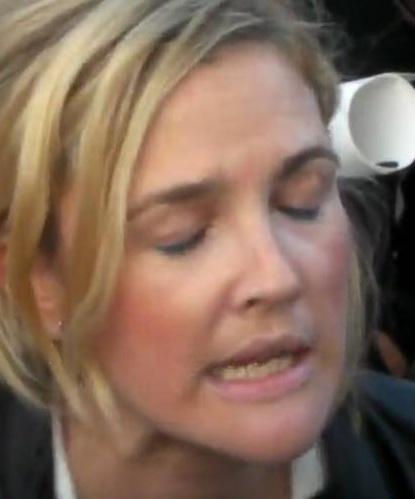}
\end{minipage}
\begin{minipage}[t]{0.89\textwidth}
\centering
\includegraphics[trim={40px 30px 0px 0px}, clip, width=\textwidth]{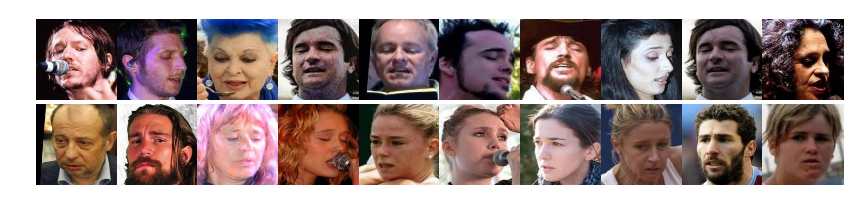}
\end{minipage}
\vspace{15pt}
\end{minipage}

\begin{minipage}[t]{\textwidth}
\centering
\begin{minipage}[t]{0.1\textwidth}
\centering
\includegraphics[width=0.85\textwidth, height=0.85\textwidth]{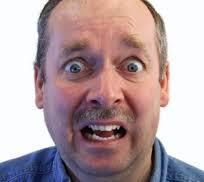}
\end{minipage}
\begin{minipage}[t]{0.89\textwidth}
\centering
\includegraphics[trim={40px 30px 0px 0px}, clip, width=\textwidth]{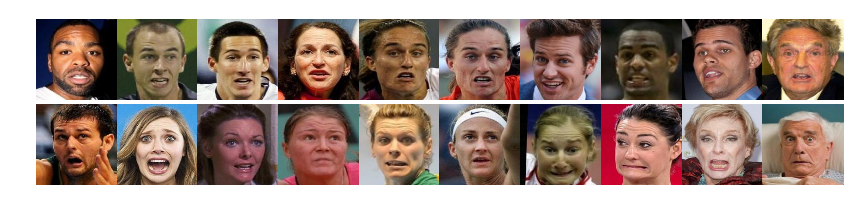}
\end{minipage}
\vspace{15pt}
\end{minipage}

\begin{minipage}[t]{\textwidth}
\centering
\begin{minipage}[t]{0.1\textwidth}
\centering
\includegraphics[width=0.85\textwidth, height=0.85\textwidth]{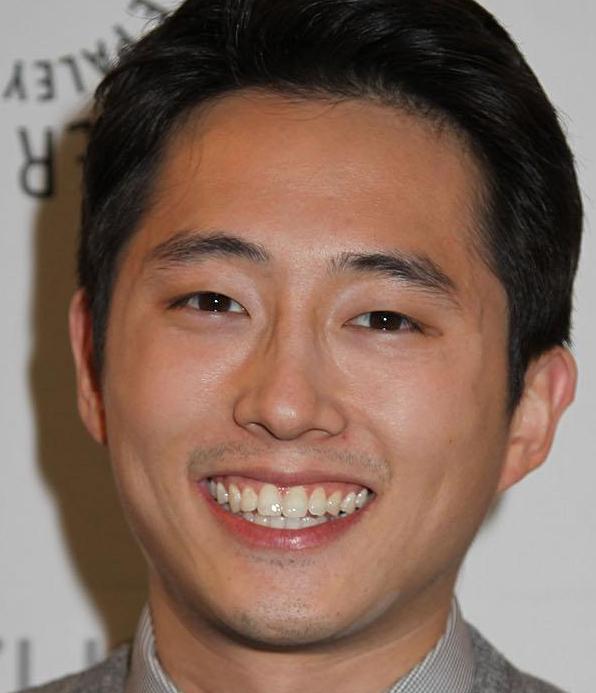}
\end{minipage}
\begin{minipage}[t]{0.89\textwidth}
\centering
\includegraphics[trim={40px 30px 0px 0px}, clip, width=\textwidth]{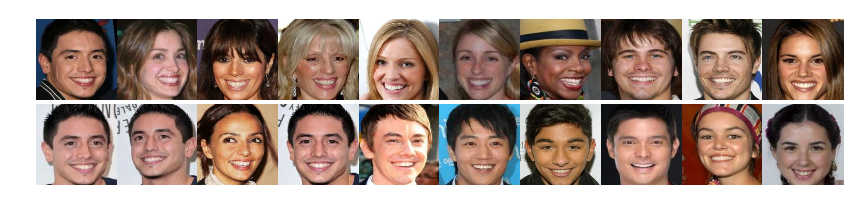}
\end{minipage}
\vspace{15pt}
\end{minipage}

\begin{minipage}[t]{\textwidth}
\centering
\begin{minipage}[t]{0.1\textwidth}
\centering
\includegraphics[width=0.85\textwidth, height=0.85\textwidth]{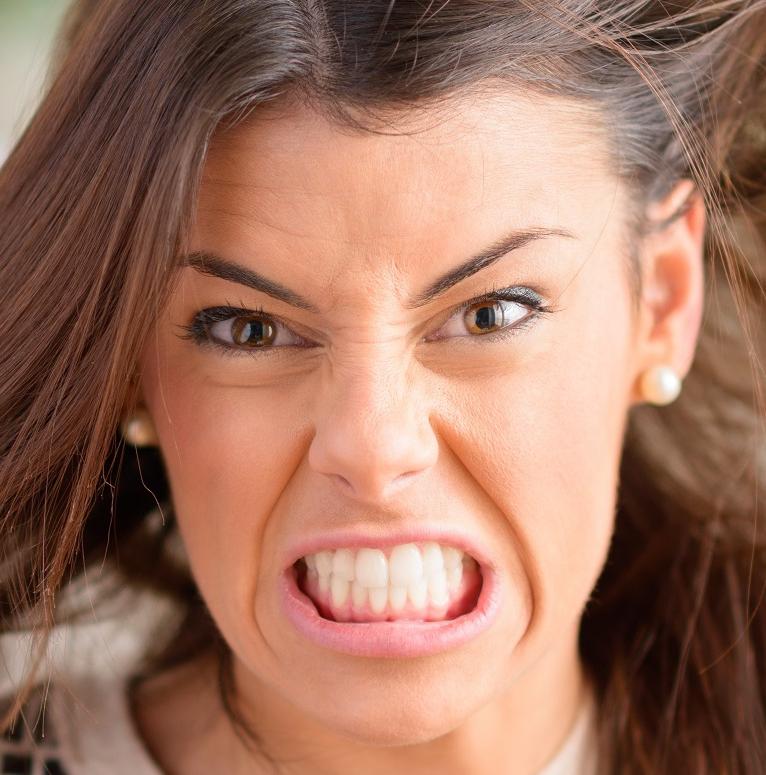}
\end{minipage}
\begin{minipage}[t]{0.89\textwidth}
\centering
\includegraphics[trim={40px 30px 0px 0px}, clip, width=\textwidth]{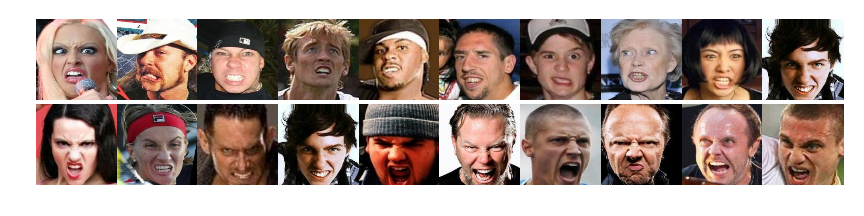}
\end{minipage}
\vspace{15pt}
\end{minipage}

\begin{minipage}[t]{\textwidth}
\centering
\begin{minipage}[t]{0.1\textwidth}
\centering
\includegraphics[width=0.85\textwidth, height=0.85\textwidth]{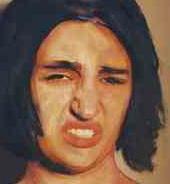}
\end{minipage}
\begin{minipage}[t]{0.89\textwidth}
\centering
\includegraphics[trim={40px 30px 0px 0px}, clip, width=\textwidth]{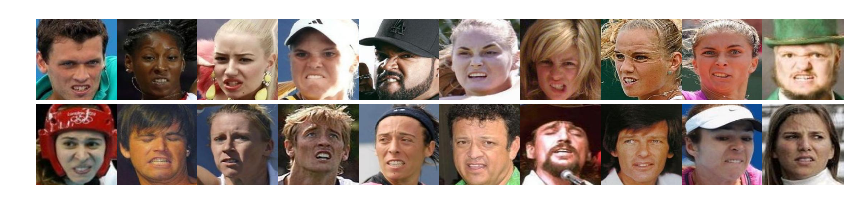}
\end{minipage}
\vspace{15pt}
\end{minipage}
\captionof{figure}{Expression retrieval results - Images on the left are the query expressions. For each query, the top row corresponds to the FECNet-16d embedding and the bottom row corresponds to the AFFNet-CL-P embedding.}
\label{fig:retrieval-last}
\end{table*}

\clearpage
\clearpage
{\small
\bibliographystyle{ieee}
\bibliography{egbib}
}
\end{document}